\pdfoutput=1

\documentclass[11pt]{article}

\usepackage[preprint]{acl}

\usepackage{times}
\usepackage{latexsym}

\usepackage[T1]{fontenc}

\usepackage[utf8]{inputenc}

\usepackage{microtype}

\usepackage{inconsolata}

\usepackage{graphicx}

\usepackage{booktabs}
\usepackage{multirow}
\usepackage{colortbl}
\usepackage{marvosym}
\usepackage{listings}
\usepackage{arydshln}
\usepackage{makecell}

\usepackage{tikz}
\usepackage[most]{tcolorbox}
\usepackage{xcolor}
\usepackage{colortbl}
\usepackage{graphicx}

\usepackage{colortbl, array, xcolor}
\usepackage{color-edits}
\usepackage{soul}
\usepackage{pifont}

\usepackage{amsmath}
\usepackage{amssymb}
\usepackage{mathtools}
\usepackage{float}

\usepackage{titletoc}
\usepackage{url}
\usepackage{hyperref}
\usepackage{natbib}

\definecolor{osgreen}{RGB}{5,156,5}
\newcommand{\myfontsize}{\footnotesize}
\newcommand{\mytextbox}[2]{%
  \tikz[baseline={([yshift=-0.5ex]current bounding box.center)}]{%
    \node[draw=#1,thick,inner sep=2pt] {\myfontsize \textbf{#2}};%
  }%
}
\definecolor{cadmiumgreen}{rgb}{0.0, 0.42, 0.24}

\definecolor{myred}{rgb}{0.7, 0.3, 0.0}
\definecolor{myblue}{rgb}{0.2, 0.3, 0.6}
\definecolor{mypurple}{rgb}{0.5, 0, 0.5}
\definecolor{mygreen}{rgb}{0.0, 0.4, 0.2}
\definecolor{mybrown}{rgb}{0.65, 0.16, 0.16}
\definecolor{mygray}{rgb}{0.5, 0.5, 0.5}

\newcommand{\corr}{\mytextbox{osgreen}{\textbf{\textcolor{osgreen}{CORRECT}}}}
\newcommand{\pcorr}{\mytextbox{myblue}{\textbf{\textcolor{myblue}{PARTIALLY CORRECT}}}}
\newcommand{\icorr}{\mytextbox{myred}{\textbf{\textcolor{myred}{INCORRECT}}}}

\definecolor{lightgreen}{rgb}{0.6, 1, 0.6}
\definecolor{lightred}{rgb}{1, 0.6, 0.6}
\definecolor{lightyellow}{rgb}{1, 1, 0.6}
\definecolor{lightblue}{rgb}{0.6, 0.8, 1}
\definecolor{lightgrey}{rgb}{0.93, 0.93, 0.93}

\newcommand{\hlgreen}[1]{{\sethlcolor{lightgreen}\hl{#1}}}
\newcommand{\hlred}[1]{{\sethlcolor{lightred}\hl{#1}}}

\newcommand{\hlblue}[1]{{\sethlcolor{lightblue}\hl{#1}}}
\newcommand{\hlgrey}[1]{{\sethlcolor{lightgrey}\hl{#1}}}

\definecolor{lakeblue}{rgb}{0.0, 0.7, 0.7}
\newcommand{\PrefRAG}{\textcolor{lakeblue}{\textbf{Pref}}\textcolor{black}{\textit{RAG}}}

\title{\textbf{\PrefRAG}: Preference-Driven Multi-Source Retrieval \\ Augmented Generation}

\author{
 Qingfei Zhao$^{1,2\dagger}$,
 Ruobing Wang$^{1,2}$,
 Yukuo Cen$^{4}$,
 Daren Zha$^{1}$,
 Shicheng Tan$^{3}$,
 Jie Tang$^{3*}$
\\
\\
 \textsuperscript{1}Institute of Information Engineering,
Chinese Academy of Sciences;\\
 \textsuperscript{2}School of Cyber Security,
University of Chinese Academy of Sciences;\\
 \textsuperscript{3}Tsinghua University;
 \quad
 \textsuperscript{4}Zhipu AI\\
 \texttt{\{zhaoqingfei,wangruobing,zhadaren\}@iie.ac.cn} \quad
 \texttt{yukuo.cen@zhipuai.cn}\\
 \texttt{tsctan@foxmail.com}
 \quad \texttt{jietang@tsinghua.edu.cn}
}

\begin{document}
\maketitle
\renewcommand{\thefootnote}{\fnsymbol{footnote}}
    \footnotetext[1]{Corresponding author
    }
\renewcommand{\thefootnote}{\fnsymbol{footnote}}
    \footnotetext[2]{Work done when QZ interned at Zhipu.AI.
    }   
\renewcommand{\thefootnote}{\arabic{footnote}}
\begin{abstract}
Retrieval-Augmented Generation (RAG) has emerged as a reliable external knowledge augmentation technique to mitigate hallucination issues and parameterized knowledge limitations in Large Language Models (LLMs).
Existing adaptive RAG (ARAG) systems excel at in-depth exploration within a single source but struggle to effectively and controllably explore different retrieval sources, as they fail to foresee their internal knowledge features.
We develop a novel multi-source ARAG system, PrefRAG, which enhances RAG by enabling in-depth and controllable exploration of diverse retrieval sources through preference-driven adaptive retrieval and self-reflection.
PrefRAG first fully explores controllable local sources in adaptive retrieval and supplements with the web when appropriate, ultimately selecting the optimal source for knowledge observation.
Subsequently, PrefRAG feeds answer quality feedback into the retrieval process, optimizing it from the generation perspective to produce higher-quality responses. Extensive experiments confirm its superiority, high retrieval efficiency, and knowledge controllability.
PrefRAG outperforms Vanilla RAG and the leading MS-ARAG by up to 25.6\% and 13.9\% respectively.
Additionally, PrefRAG trained with DPO achieves higher performance. The code and data are available at \url{https://github.com/QingFei1/PrefRAG.git}.
\end{abstract}

\section{Introduction}
In the question answering (QA) task~(\citealp{DBLP:journals/tacl/KwiatkowskiPRCP19};
~\citealp{DBLP:conf/emnlp/RajpurkarZLL16}), even the leading Large Language Models (LLMs) (\citealp{DBLP:journals/corr/abs-2303-08774};~\citealp{DBLP:journals/corr/abs-2406-12793};~\citealp{DBLP:journals/corr/abs-2302-13971}) are restricted by the scope of their parametric knowledge and struggle with hallucination~\citep{DBLP:conf/cikm/ChenFYWFL0LX23} and insufficient knowledge~\citep{DBLP:conf/icml/KandpalDRWR23}.
\begin{figure}[htbp]
\includegraphics[width=\columnwidth]{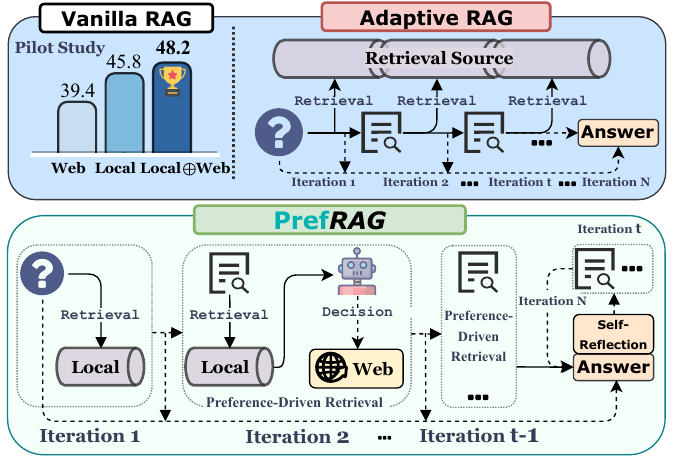}
\caption{\textbf{Comparison of Different Methods.}
Single-source adaptive RAG enables in-depth exploration but cannot integrate cross-source knowledge. PrefRAG addresses this limitation by enabling efficient and adaptive exploration of different retrieval resources.
}
\label{fig:PrefRAG-Figure1}
\end{figure}
Retrieval-Augmented Generation (RAG)~\citep{DBLP:conf/nips/LewisPPPKGKLYR020} serves as a powerful technique that mitigates these challenges by supplementing external knowledge with a non-parametric form, generating high-quality and reliable answers.
Mainstream retrieval sources for RAG typically include local retrieval sources, e.g., Wikipedia corpus~\citep{DBLP:journals/jmlr/IzacardLLHPSDJRG23} or web retrieval sources, e.g., Bing, each with distinct data characteristics~\citep{DBLP:journals/jasis/Williams00}.
Generally, local retrieval sources are carefully curated, highly structured, and offer greater control and security due to their on-premise storage.
In contrast, web-based retrieval sources provide large-scale, diverse, and real-time information but are inherently less controllable.
These differences indicate that each retrieval source has its own advantages and limitations.
\textbf{A pilot study} conducted on a multi-hop dataset~\citep{DBLP:conf/coling/HoNSA20}, as illustrated in Fig.~\ref{fig:PrefRAG-Figure1}, reveals that knowledge from local and web sources can be mutually reinforcing, leading to enhanced performance.

However, existing RAG remain underdeveloped in their ability to effectively and controllably leverage multiple retrieval sources with distinct characteristics.
\textbf{As depicted in Fig.~\ref{fig:PrefRAG-Figure1}}, Adaptive RAG (ARAG)~(\citealp{DBLP:conf/emnlp/JiangXGSLDYCN23};~\citealp{DBLP:conf/naacl/JeongBCHP24}) typically focus on exploring a single retrieval source (either local or web) in depth, overlooking the complementary contributions of multiple sources.
Recently, an LLM-based agent paradigm, ReAct~\citep{DBLP:conf/iclr/YaoZYDSN023} can be instantiated as Multi-Source ARAG (MS-ARAG) and allow retrieval from multiple sources throughout the iterative process.
However, ReAct struggles to foresee the data characteristics in different retrieval sources before retrieval.
Its source selection decision relies on retrieval source descriptions and the model’s internal parameterized representation, which may fail to align with the real retrieval demands.
Another direct strategy for leveraging diverse sources is concatenating knowledge from different sources.
This strategy risks direct exposure of problematic web content to the LLM, potentially generating undesirable outputs and requiring more retrieval counts.

\begin{figure*}[htbp]
\includegraphics[width=\textwidth]{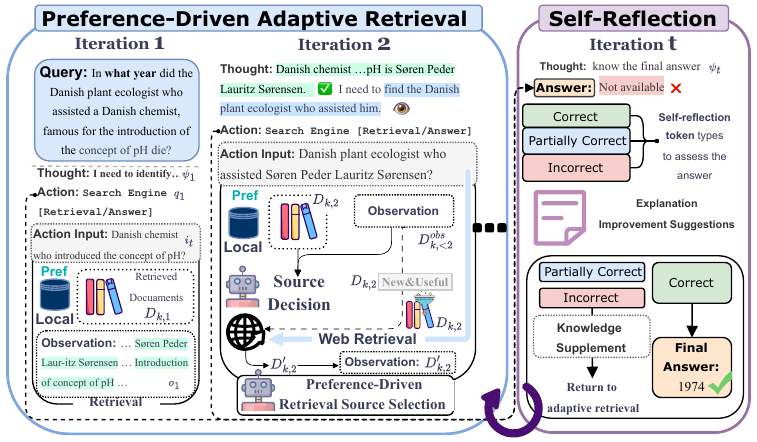}
\caption{\textbf{Overview of PrefRAG.} PrefRAG comprises a preference-driven adaptive retrieval process (\textit{left}) and a self-reflection process (\textit{right}).
}
\label{fig:overall}
\end{figure*}

To bridge these gaps, we develop PrefRAG, a novel MS-ARAG system designed for efficient, controlled, and adaptive exploration of retrieval sources with diverse characteristics.
As illustrated in Fig.~\ref{fig:overall}, PrefRAG consists of two core processes: preference-driven adaptive retrieval (Pref-AR) and self-reflection.
During the Pref-AR process, the LLM decides whether to retrieve and what to retrieve based on the original query and accumulated context, enabling adaptive retrieval.
Once a retrieval action is determined, we retrieve the preset preferred source (e.g., the local source) and then guide the LLM to analyze the retrieved knowledge before deciding whether to switch to another source (e.g., the web source).
This enables the system to conduct in-depth knowledge analysis and make well-considered retrieval source decisions.
Moreover, such an orderly retrieval process transitioning from the relatively controlled local source to the web source helps minimize the risk of exposing the LLM to uncontrolled knowledge from the web when local retrieval suffices.
During the self-reflection process, the LLM assesses the reliability of responses and provides specific improvement suggestions through self-feedback~(\citealp{DBLP:conf/nips/MadaanTGHGW0DPY23};~\citealp{DBLP:conf/nips/ShinnCGNY23}), thereby guiding subsequent retrieval and reasoning processes to enhance the final response quality.

To summarize, our main contributions are as follows: \textbf{1)} We develop a novel MS-ARAG system with preference-driven adaptive retrieval and self-reflection mechanisms.
The system leverages preference constraints to guide the RAG system in selecting appropriate retrieval sources and refines subsequent retrieval through self-reflection, enabling deep and controllable knowledge utilization from diverse retrieval sources to generate high-quality answers.
\textbf{2)} We propose an automated pipeline for constructing preference-driven retrieval training data, which generates high-quality data for Direct Preference Optimization (DPO)~\citep{DBLP:conf/nips/RafailovSMMEF23} fine-tuning, further enhancing the system's capability. \textbf{3)} Extensive empirical studies conducted on four datasets demonstrate the effectiveness of PrefRAG.
Experimental results show that our method significantly outperforms Vanilla RAG (by up to 25.6\%) and the leading MS-RAG (by up to 13.9\%) while maintaining high retrieval efficiency.
In real-world applications, we further validate the superior performance of PrefRAG in controllable knowledge retrieval.

\section{Related Work}
\noindent\textbf{Knowledge Source Exploration for RAG.}
In the era of LLM, RAG~(\citealp{DBLP:conf/nips/LewisPPPKGKLYR020};~\citealp{DBLP:conf/icml/GuuLTPC20}) builds on the versatile LLM as a foundation and serves as a bridge between external knowledge and the model's internal parameterized knowledge by following the "\textit{Retriever-and-Reader}" paradigm~(\citealp{DBLP:conf/acl/ChenFWB17};~\citealp{DBLP:conf/iclr/DasDZM19}).
For various downstream tasks~(\citealp{DBLP:journals/corr/abs-2101-00774};~\citealp{DBLP:conf/iclr/Zhou0XJN23};~\citealp{cai-etal-2019-skeleton}), RAG systems retrieve accessible sources as comprehensively as possible to enhance generation, especially for knowledge-intensive question answering task~\citep{DBLP:journals/corr/abs-2212-14024}.
In terms of the manner of retrieval sources, recent advanced RAG research can be divided into two categories.
One line of study conducts in-depth exploration within a single retrieval source, referred to as Single-Source RAG (SS-RAG). It primarily includes multi-step RAG methods~(\citealp{DBLP:conf/acl/TrivediBKS23};~\citealp{DBLP:journals/tacl/RamLDMSLS23};~\citealp{DBLP:conf/icml/BorgeaudMHCRM0L22}) that use subqueries for iterative retrieval and ARAG methods~(\citealp{DBLP:conf/iclr/YaoZYDSN023};~\citealp{DBLP:conf/iclr/AsaiWWSH24};~\citealp{dhole2025retrieveretrieveuncertaintydetection}) that flexibly determine "\textit{when and what to retrieve}" for a more adaptive and in-depth retrieval process.
For Single-Source ARAG (SS-ARAG), the limitation of a single retrieval source imposes an upper bound on the capability of the RAG system.
Another line of research focuses on Multi-Source RAG (MS-RAG). CRAG~\citep{DBLP:journals/corr/abs-2401-15884} uses the web as a backup retrieval source, while ReAct, an agent framework, can be instantiated to achieve basic MS-ARAG. However, it cannot foresee the features of different retrieval sources and heavily relies on their descriptions for selection, leading to low-quality and unstable multi-source retrieval.
Therefore, PrefRAG aims to achieve adaptive retrieval while ensuring a stable selection of the most suitable retrieval source during iteration.

\noindent \textbf{Fine-Tuning for RAG.}
In traditional RAG, fine-tuning methods are widely employed to enhance the retriever and generator~(\citealp{DBLP:conf/iclr/Lin0CSL00KSLZY24};~\citealp{DBLP:conf/acl/KeK00MB24}). Beyond this, modular RAG systems integrate a series of LLM-based components~\citep{DBLP:journals/corr/abs-2312-10997}. Fine-tuning helps models better follow complex instructions within these components~\citep{DBLP:conf/emnlp/HeZHLX24}, improving RAG systems' performance and task adaptability~(\citealp{DBLP:conf/iclr/AsaiWWSH24};~\citealp{DBLP:journals/corr/abs-2403-10131};~\citealp{DBLP:conf/naacl/JeongBCHP24}). 
Classic supervised fine-tuning strategy (SFT) trains only on positive samples. While DPO as a more direct reinforcement learning
fine-tuning (RLFT) method, leverages positive-negative sample pairs to effectively and efficiently strengthen LLMs' ability to follow complex instructions.
Under the multi-source setting, our work thus employs DPO to enhance the model's ability to follow the retrieval selection instruction to select the optimal retrieval source during adaptive retrieval.

\section{\PrefRAG}

\subsection{Task Definition and Overview}
\label{subsec:Task Definition and Overview}
Following the retrieval-and-generation paradigm of Vanilla RAG, PrefRAG leverages two different types of mainstream retrieval sources with distinct characteristics, i.e., local corpus $S_L$ and web browser $S_W$, denoted as $ \left \{S_L,S_W \right \} \in S $.
Notably, PrefRAG can handle more than two retrieval sources, as detailed in Appendix~\ref{Appendixsec-E: More Retrieval Sources}.

We present an overview of PrefRAG in Fig.~\ref{fig:overall}. Given an original query $q$, PrefRAG performs \textbf{preference-driven adaptive retrieval} process and \textbf{self-reflection} process.
During preference-driven adaptive retrieval, PrefRAG iteratively yields reasoning thought $\psi \in \Psi$, preference-driven retrieval decision (including actions $a_t\in \mathcal{A}$, action inputs $q_t\in Q$ as subqueries, and retrieval selection decision $S_{\text{Dec}}$), then construct retrieval source observations $o_t\in \mathcal{O}$ based on a preset retrieval preference for $S_L$.
Answer generation serves as the stopping criterion for this adaptive retrieval process. 
We define the iteration process as $\{\tau_t\}_{t=1}^{n},\,\,n \in \mathbb{N}^{+}$.
Each iteration $\tau_{t}$ starts with the thought generation process.

During self-reflection, PrefRAG outputs a self-reflection token for the answer $\alpha$, along with explanations and improvement suggestions.
if a negative self-reflection token is triggered, it re-engages the adaptive retrieval process, repeating iterations until a self-revised final answer $\alpha$ is generated.

\subsection{Preference-Driven Adaptive Retrieval}
\label{subsec:Preference-Driven Adaptive Retrieval}
Constructing high-performance MS-ARAG systems faces several challenges. \textbf{For adaptive retrieval}, systems need to decompose questions, plan problem-solving paths, and determine retrieval timing based on existing reasoning. \textbf{For multi-source retrieval}, one potential risk is that systems cannot foresee source characteristics relying on brief descriptions. Systems also tend to exclude previously low-quality sources, limiting further exploration.

To this end, we propose a preference-driven adaptive retrieval process, which consists of three subprocesses: reasoning thought, preference-driven retrieval decision, and source observation.

\noindent\textbf{Reasoning Thought.} The LLM generates a free-form reasoning thought $\psi_{1}$ from the original query $q$.  The reasoning thought involves decomposing the query and outlining a solution path, guiding subsequent retrieval decisions. In later iterations, the reasoning thought $\psi_{t}$ is derived from both $q$ and the accumulated context $c_{t-1}$:

\begin{equation}
\psi_{t} \sim \text{LLM}_{\text{AR}}(\text{Instruct}_{\text{AR}}, q \Vert c_{t-1} )
    \label{eq4.1:Thought}
\end{equation} 
Specifically, the $c_{t-1}$ represents the accumulated context from previous iterations $\tau_{<t}$, encompassing retrieval actions $\{a_i\}_{i=1}^{t-1}$ and their corresponding action inputs $\{q_i\}_{i=1}^{t-1}$, retrieved source observations $\{o_i\}_{i=1}^{t-1}$.
The $\text{Instrut}_{\text{AR}}$ represents the prompt for generating thoughts (cf. Appendix~\ref{Appendixsec-B.2:Prompt}). The $\text{LLM}_{\text{AR}}$ indicates the LLM used in the process of generating thought $\psi_{t}$.

\noindent\textbf{Preference-Driven Retrieval Decision.}
After generating a reasoning thought $\psi_{t}$, we direct the LLM in developing a two-stage retrieval decision by leveraging the cues in the $\psi_{t}$ and the $c_{t}$.
The two-stage retrieval decision includes the "\textit{Retrieve-or-Generate}" and the "\textit{Retrieval Source Selection}" decision stage.
In the \textit{Retrieve-or-Generate} stage, the system determines whether to proceed to adaptive retrieval or answer generation. If choosing to continue retrieval, the LLM outputs "\texttt{Search\_Engine}" as the [\textit{Action}] token and formulates a subquery $q_{t}$ as the [\textit{Action Input}]. Alternatively, if the LLM outputs an answer $\alpha$, the RAG system enters a self-reflection process (\S~\ref{subsec:Self-reflection}).

In the \textit{retrieval source selection}  stage, we implement a "preference-first retrieval with conditional switching" strategy.  The RAG system initially prioritizes retrieving from a curated local source $S_{L}$. Using the subquery $q_{t}$ from the [\textit{Action Input}], the retriever $\mathcal{R}$ obtains top-$k$ documents $D_{k,t}=\left \{ d_{1},d_{2},\cdots,d_{k} \right \}$  from $S_{L}$.
Subsequently, we instruct the LLM to compare the newly retrieved documents $D_{k,t}$ in $\tau_{t}$ with the previously observed documents $D^{obs}_{k,<t}$ to determine whether to switch to the web retrieval source.
The $D^{obs}_{k,<t}$ represents all documents $o_{1},o_{2},\dots,o_{t-1}$, accumulated in context $c_{t}$  from previous iterations $\tau_{<t}$. This comparison process enables the system to continuously perceive knowledge feedback from retrieval sources, thereby improving the LLM's follow-up inference.

\begin{equation}
D^{obs}_{k,<t} \coloneqq \left\{o_{1},o_{2},\dots,o_{t-1}\right\}\subsetneq c_{t}
    \label{eq4.2.2:Decision}
\end{equation}
Equation~(\ref{eq4.2.2:Decision}) clearly describes relationships among these variables.
To sum up, here is the mathematical expression of the comparison process:
\begin{equation}
\begin{aligned}
S_{\text{Dec}} & \sim \text{LLM}_{\text{Sel}}(\text{Instruct}_\text{Sel},q\Vert D_{k,t}\Vert D^{obs}_{k,<t})\\
S_{\text{Dec}} & = \begin{cases}
\operatorname{analysis} \mapsto \text{CoT}_{\text{Dec}}, \\
\operatorname{status} \mapsto V_{\text{Dec}}
\end{cases}\\
\end{aligned}
\label{eq4.2.3:Selection}
\end{equation}
The $\text{LLM}_{\text{Sel}}$ and $\text{Instruct}_{\text{Sel}}$ refer to the model and prompt used for this comparison process (cf. Appendix~\ref{Appendixsec-B.2:Prompt}).
The $S_{\text{Dec}}$ denotes the comparison result. Specifically, the $\text{LLM}_{\text{Sel}}$ first outputs a Chain-of-Thought analysis ($\text{CoT}_{\text{Dec}}$), which explicitly guides the subsequent generation of the status value ($V_{\text{Dec}}$), thereby enhancing the accuracy of the comparison result.
A status value of \texttt{True} indicates that the local retrieval source sufficiently satisfies the knowledge requirements of the $q$ in the current iteration $\tau_{t}$, making additional retrieval from the web unnecessary. Conversely, a status value of \texttt{False} signifies switching to the web retrieval source, then retrieving the top-$k$ documents from the web.

\noindent\textbf{Retrieval Source Observation.}
For the RAG system to adaptively refine retrieval decisions, the LLM should account for feedback from retrieved knowledge, thereby improving subsequent retrieval decisions through in-context learning.
In iteration $\tau_{t}$, if $V_{Dec}=\texttt{True}$, we use the $D_{k,t}$ from the local source as the content of  $o_{t}$; if $V_{Dec}=\texttt{False}$, we use only the $D'_{k,t}$ from the web source.

\subsection{Self-Reflection}
\label{subsec:Self-reflection}
Existing ARAG systems may generate erroneous final answers in complex tasks in some cases due to low-quality retrieval.
Therefore, it is essential to refine the retrieval strategy based on feedback from the final answer.
We develop a self-reflection process to critically assess responses and further explore retrieval sources when necessary.

\noindent\textbf{Answer Assessment.}
After the LLM generates an answer $\alpha$, we instruct the LLM to produce a self-reflection token accompanied by a brief explanation.
Specifically, this self-reflection token assesses the quality of the generated $\alpha$ informed by $c_{t-1}$.
To simplify the evaluation task, we classify the assessment results into three discrete classes: {\corr}, {\pcorr}, and {\icorr}.  When the LLM outputs "CORRECT", the RAG system considers the current answer as final.
For negative assessments ("PARTIALLY CORRECT"/"INCORRECT"), the LLM first generates the explanation and improvement suggestion to highlight aspects of the answer that need refinement or correction, then triggers further retrieval.

\noindent\textbf{Multi-Source Knowledge Supplement.}
When the model outputs negative self-reflection tokens, we concurrently use the $q$ to retrieve from both local $S_{L}$ and web sources $S_{W}$. Next, we incorporate all documents retrieved from these sources into the [\textit{Observation}] as supplementary knowledge. 
The context of current iteration, including thought $\psi_{t}$, answer $\alpha$, self-reflection process, is added to $c_{t-1}$ as $c_{t}$.
Subsequently, the RAG system re-enters the preference-driven adaptive retrieval process (\S~\ref{subsec:Preference-Driven Adaptive Retrieval}).
Such a knowledge supplementation strategy allows the system to leverage the most relevant information from multiple sources related to $q$, enhancing the quality of subsequent $\alpha$, especially when we know the current answer quality is low.

\noindent\textbf{Iteration Termination Condition.}
We establish two iteration termination conditions for the PrefRAG. The system terminates and regards the current answer as final when the self-reflection label of the $\alpha$ is \corr.
Alternatively, it stops when the preference-driven adaptive retrieval process reaches the preset maximum number of iterations, irrespective of the type of self-reflection token.

\subsection{DPO Data Construction}
\label{subsec:DPO Data Construction}
We propose an automated pipeline for constructing preference-driven retrieval source selection data for training.
Due to the high cost of human annotation, 
we use GLM4-Plus to generate retrieval source selection labels to simulate human preferences.
The input $x$ in the training data consists of the instruction template $\text{Instruct}_{\text{sel}}$, query $q$, retrieved documents $D_{k,t}$, and previously observed documents $D^{obs}_{k,<t}$.
Using this input, GLM4-9B-chat generates multiple candidate responses, and then we use GLM4-Plus to identify positive $y^{+}$ and negative $y^{-}$ response pairs.
Ultimately, our training dataset $\mathcal{D}$ comprises 4000 samples, with each sample represented as $\left\{x, y^{+}, y^{-}\right\} \sim \mathcal{D}$ \textbf{(more details on data construction in Appendix~\ref{Appendixsec-C: DPO Data Construction Details})}.

\subsection{Training for Alignment (DPO)}
\label{subsec:Training for Alignment (DPO)}
During training, we employ DPO, a method that straightforwardly trains the aligned model, and the optimization objective is:
\begin{multline}
\mathcal{L}(M^{\theta}_{\text{Sel}};M^{ref}_{\text{Sel}})=-\mathbb{E}_{\left \{ x,y^{+},y^{-} \right \}\sim \mathcal{D}} [log\sigma \\ [  \beta log\frac{M^{\theta}_{\text{Sel}}(y^{+}|x)}{M^{ref}_{\text{Sel}}(y^{+}|x)}-\beta log\frac{M^{\theta}_{\text{Sel}}(y^{-}|x)}{M^{ref}_{\text{Sel}}(y^{-}|x)}]]
\end{multline}
where $M^{\theta}_{\text{Sel}}$ stands for the DPO-trained model, and $M^{ref}_{\text{Sel}}$ serves as a reference model initialized from the built-in model $\text{LLM}_{\text{Sel}}$ of the retrieval source selection process.
Additionally, we conduct full parameter fine-tuning on 8×A100 GPUs (80GB each), with $\beta$ = 0.1, a batch size of 8, and a learning rate of 5e-7, training the model for one epoch.

\section{Experimental Setup}
\subsection{Datasets \& Metrics \& Retrieval Settings}
\noindent \textbf{Datasets} \quad Following previous work~ (\citealp{DBLP:conf/iclr/YaoZYDSN023};~\citealp{DBLP:conf/acl/TrivediBKS23};~\citealp{DBLP:conf/acl/Xiong0LZ24}), we evaluate on both open-domain and domain-specific QA datasets. For open-domain QA, we select three challenging multi-hop datasets: HotpotQA~\citep{DBLP:conf/emnlp/Yang0ZBCSM18}, 2WikiMultiHopQA (2WikiMQA)~\citep{DBLP:conf/coling/HoNSA20}, and MuSiQue~\citep{DBLP:journals/tacl/TrivediBKS22}. For domain-specific QA, we select BioASQ-Y/N~(\citealp{DBLP:journals/bmcbi/TsatsaronisBMPZ15}; \citealp{krithara2023bioasq}), which requires \texttt{Yes/No} answers based on biomedical knowledge \textbf{(more details in Appendix~\ref{Appendixsec-B.1:Datasets})}.

\noindent\textbf{Evaluation Metrics} \quad We adopt Exact Match (EM) and F1-score (F1) for multi-hop QA ~\citep{DBLP:conf/emnlp/JiangXGSLDYCN23}, and Accuracy (Acc.) for both multi-hop ~\citep{vu2020ava} and biomedical QA~\citep{DBLP:conf/acl/Xiong0LZ24}.

\begin{table*}[ht]
\renewcommand{\arraystretch}{0.95}
\setlength{\tabcolsep}{3.3pt}
    \fontsize{8}{8.5}\selectfont
  \centering
  
    \begin{tabular}{lccccccccccccc}
    \toprule
          \multirow{2}[2]{*}{\textbf{Methods \& LLMs}} & \multicolumn{4}{c}{\bf HotpotQA} & \multicolumn{4}{c}{\bf 2WikiMQA} & \multicolumn{4}{c}{\bf MuSiQue} & {\bf BioASQ-Y/N}\\
\cmidrule(lr){2-5}\cmidrule(lr){6-9}\cmidrule(lr){10-13}\cmidrule(lr){14-14}        & Acc. & F1 & EM & Avg. & Acc. & F1 & EM & Avg. & Acc. & F1 & EM & Avg. & Acc. \\
    \hline
    \rowcolor[rgb]{ .851,  .851,  .851} \multicolumn{14}{c}{\textbf{\# Baselines without Retrieval (NoR) \#}} \\
    \multicolumn{14}{c}{\textit{Open-source LLMs}} \\
    Llama3.1-8B-Instruct & 22.6 & 28.7 & 23.0 & 24.8 & 27.4 & 30.7 & 26.4 & 28.2 & 3.6 & 9.4 & 3.2 & 5.4 & 77.8 \\
GLM4-9B-chat & 18.4 & 23.5 & 17.4 & 19.8 & 25.6 & 29.6 & 25.0 & 26.7 & 3.0 & 8.8 & 2.6 & 4.8 & 74.0 \\
\hdashline
    \multicolumn{14}{c}{\textit{Proprietary LLMs}} \\
    GPT-4o-mini & 29.8 & 38.4 & 28.6 & 32.3 & 29.2 & 32.6 & 26.6 & 29.5 & 7.6 & 15.4 & 5.0 & 9.3 & 86.6 \\
    GLM4-Plus & 30.2 & 38.3 & 29.8 & 32.8 & 30.4 & 35.2 & 29.6 & 31.7 & 8.2 & 15.8 & 7.2 & 10.4 & 81.8  \\
    \hline
    \rowcolor[rgb]{ .851,  .851,  .851} \multicolumn{14}{c}{\textbf{\# Vanilla RAG \#}} \\
    \multicolumn{14}{c}{\textit{Only local retrieval source (Vanilla \textit{w/ LR})}} \\
    Llama3.1-8B-Instruct  & 36.4 & 45.6 & 34.4 & 38.8 & 31.2 & 35.4 & 30.2 & 32.3 & 6.4 & 12.2 & 5.6 & 8.1 & 85.8 \\
    GLM4-9B-chat & 34.8 & 44.4 & 34.2 & 37.8 & 34.4 & 38.8 & 33.8 & 35.7 & 8.2 & 15.0 & 7.0 & 10.1 & 87.2 \\
    GPT-4o-mini  & 45.0 & 53.8 & 41.2 & 46.7 & 40.2 & 44.2 & 38.6 & 41.0 & 11.2 & 19.2 & 8.8 & 13.1 & 89.6 \\
    GLM4-Plus  & 46.4 & 56.7 & 45.8 & 49.6 & 45.6 & 48.9 & 43.0 & 45.8 & 15.4 & 23.5 & 13.8 & 17.6 & 89.8 \\
    \hdashline
    \multicolumn{14}{c}{\textit{Concatenating both local and web retrieval source (Vanilla${_{\text{ Mix}}}$ w/ LR $\oplus $ WR) }} \\
    Llama3.1-8B-Instruct  & 41.6 & 53.9 & 41.2 & 45.6 & 35.4 & 39.3 & 32.6 & 35.8 & 9.0 & 16.0 & 8.0 & 11.0 & 89.6 \\
    GLM4-9B-chat & 40.8 & 51.3 & 39.0 & 43.7 & 38.8 & 43.7 & 37.4 & 40.0 & 9.0 & 16.7 & 8.4 & 11.4 & 91.0 \\
    GPT-4o-mini  & 47.4 & 58.0 & 44.6 & 50.0 & 45.8 & 49.1 & 40.6 & 45.2 & 13.2 & 21.3 & 11.4 & 15.3 & 92.2 \\
    GLM4-Plus  & 49.6 & 61.1 & 48.4 & 53.0 & 48.4 & 51.7 & 44.6 & 48.2 & 13.6 & 23.9 & 13.2 & 16.9 & \underline{93.6} \\
    \hline
    \rowcolor[rgb]{ .851,  .851,  .851} \multicolumn{14}{c}{\textbf{\# Single-Source ARAG (SS-ARAG) \#}} \\
    FLARE $_{\text{GLM4-Plus}}$ & 46.4 & 51.8 & 41.8 & 46.7 & 49.4 & 45.9 & 37.8 & 44.4 & 16.6 & 21.9 & 14.4 & 17.6 & 77.2 \\
    Self-RAG $_{\text{GLM4-Plus}}$ & 45.0 & 54.5 & 43.6 & 47.7 & 32.4 & 36.7 & 30.2 & 33.1 & 15.4 & 24.3 & 13.2 & 17.6 & 82.8 \\
    \rowcolor[rgb]{ .851,  .851,  .851} \multicolumn{14}{c}{\textbf{\# Multi-Source RAG (MS-RAG) \#}} \\
CRAG $_{\text{GLM4-Plus}}$ & 41.8 & 50.1 & 37.8 & 43.2 & 35.2 & 37.6 & 29.0 & 33.9 & 11.6 & 17.4 & 8.8 & 12.6 & 89.0 \\
ReAct \textit{w/ LR $\&$ WR} $_{\text{GLM4-Plus}}$ & 50.0 & 59.7 & 46.2 & 52.0 & 64.2 & 63.8 & 51.8 & 59.9 & 23.2 & 30.6 & 18.4 & 24.1 & 91.8 \\
ReAct${_{\textit{Mix}}}$ \textit{w/ LR $\oplus$ WR} $_{\text{GLM4-Plus}}$ & 56.6 & \underline{67.0} & \underline{53.6} & \underline{59.1} & 73.8 & 70.5 & 59.0 & 67.8 & 25.8 & 33.3 & \underline{21.2} & 26.8 & 93.2 \\
    \hline
    \rowcolor[rgb]{0.85, 0.985, 0.985} \multicolumn{14}{c}{\textbf{\# Ours \#}} \\
PrefRAG $_{\text{Llama3.1-8B-Instruct}}$ & 42.0 & 51.1 & 38.8 & 44.0 & 42.0 & 43.2 & 35.8 & 40.3 & 15.4 & 21.0 & 12.8 & 16.4 &  89.6 \\
PrefRAG $_{\text{GLM4-9B-chat}}$ & 45.4 & 56.3 & 42.2 & 48.0 & 55.0 & 53.7 & 42.0 & 50.2 & 23.0 & 29.4 & 20.0 & 24.1 & 87.6 \\
PrefRAG-DPO $_{\text{GLM4-9B-chat}}$ & 51.4 & 57.0 & 45.0 & 51.1 & 57.0 & 56.0 & 45.2 & 52.7 & 24.2 & 30.0 & 20.2 & 24.8 & 89.6 \\
PrefRAG $_{\text{GPT-4o-mini}}$ & \underline{58.6} & 66.0 & 50.4 & 56.6 & \underline{76.2} & \underline{72.1} & \underline{59.4} & \underline{69.2} & \underline{28.2} & \underline{34.3} & \underline{21.2} & \underline{27.9} & 92.8 \\
PrefRAG $_{\text{GLM4-Plus}}$ & \textbf{59.0} & \textbf{68.4} & \textbf{55.0} & \textbf{60.8} & \textbf{79.6} & \textbf{76.7} & \textbf{65.2} & \textbf{73.8} & \textbf{32.2} & \textbf{39.4} & \textbf{27.4} & \textbf{33.0} & \textbf{94.0} \\
$\Delta_{\text{ GLM4-Plus}\rightarrow \text{Vanilla}\,\,\textit{w/ LR}}$ & 12.6$\uparrow$ & 11.7$\uparrow$ & 9.2$\uparrow$ & 11.2$\uparrow$ & 34.0$\uparrow$ & 27.8$\uparrow$ & 22.2$\uparrow$ & 28.0$\uparrow$ & 16.8$\uparrow$ & 15.9$\uparrow$ & 13.6$\uparrow$ & 15.4$\uparrow$ & 4.2$\uparrow$ \\
$\Delta_{\text{ GLM4-Plus}\rightarrow \text{Vanilla${_{\textit{Mix}}}$}\,\,\textit{w/ LR $\oplus $ WR}}$ & 9.4$\uparrow$ & 7.3$\uparrow$ & 6.6$\uparrow$ & 7.8$\uparrow$ & 31.2$\uparrow$ & 25.0$\uparrow$ & 20.6$\uparrow$ & 25.6$\uparrow$ & 18.6$\uparrow$ & 15.5$\uparrow$ & 14.2$\uparrow$ & 16.1$\uparrow$ & 0.4$\uparrow$ \\
    \bottomrule
    \end{tabular}%
      \caption{\textbf{Results (\%) of overall performance.} "\textbf{Bold}" and "\underline{Underlined}" denote the highest absolute values and second highest values, respectively. "\textbf{$\Delta$}" represents the increase compared to Vanilla. "\textbf{w/ LR}" denotes utilizing only local sources. "\textbf{w/ LR $\oplus$ WR}" denotes concatenating both local and web retrieval sources. "\textbf{w/ LR $\&$ WR}" denotes selecting either the local or web retrieval source at each iteration. The "Avg." denotes the arithmetic mean.}
  \label{tab:Overall Performance}%
\end{table*}%
\noindent\textbf{Retrieval Settings}\quad 
For local retrieval, we employ the corpus version released by~\citeauthor{DBLP:conf/acl/TrivediBKS23} for multi-hop QA and PubMed\footnotemark[1]~\citep{DBLP:conf/acl/Xiong0LZ24} for biomedical QA.
\footnotetext[1]{\url{https://pubmed.ncbi.nlm.nih.gov/}}
Across all datasets in local retrieval, BM25 implemented in Elasticsearch serves as the sparse retriever, while bge-large-en-v1.5\footnotemark[2] is used as the dense retriever.
\footnotetext[2]{\href{https://huggingface.co/BAAI/bge-large-en-v1.5}{\texttt{https://huggingface.co/BAAI/\\bge-large-en-v1.5}}}
For web retrieval, we adopt a public and accessible web search API, DuckDuckGo\footnotemark[3], to retrieve information from the large-scale web source.
Additionally, we experiment with different numbers of retrieved passages (more results in Appendix~\ref{Appendixsec-A.5: Different top-k Values and Retrievers}), top-$k\in \{3, 5, 7\}$, with a default value of 5.
\footnotetext[3]{\url{https://duckduckgo.com/}}

\subsection{Baselines \& LLMs}
\noindent\textbf{Baselines}\quad
We compare PrefRAG with four categories of baselines.
\textbf{No Retrieval (NoR)} refers to feeding the query directly into the LLM to output answers without
retrieval.
\textbf{Vanilla RAG (Vanilla)} represents the standard RAG, which executes a one-time retrieval and feeds the retrieved context, along with the original query, into the LLM to generate answers.
\textbf{Single-Source ARAG (SS-ARAG)} adaptively explores a single retrieval source (e.g., only local retrieval), including recent mainstream methods such as Self-RAG~\citep{DBLP:conf/iclr/AsaiWWSH24} and FLARE~\citep{DBLP:conf/emnlp/JiangXGSLDYCN23}.
\textbf{Multi-Source RAG (MS-RAG)} allows multiple retrieval sources for knowledge augmentation.
Among them, CRAG performs single-time retrieval and uses web search only at the final stage as a complement. ReAct is a classic agent framework that can be instantiated as an ARAG system.

\noindent\textbf{LLMs}\quad We conduct experiments based on five built-in LLMs, including Llama3.1-8B~\citep{DBLP:journals/corr/abs-2407-21783}, GLM4-9B, Llama3.1-70B, GPT-4o-mini~\citep{DBLP:journals/corr/abs-2410-21276} 
and GLM4-Plus~\citep{DBLP:journals/corr/abs-2406-12793}.
Our DPO training is performed on the open-source GLM4-9B model.

\subsection{Implementation Details}
To accelerate model inference, we deploy all locally hosted open-source models using the vLLM~\citep{DBLP:conf/sosp/KwonLZ0ZY0ZS23} inference acceleration toolkit.
During inference, we set the temperature to 0.1 across all models to reduce uncertainty and align answer formats in prompts across all baselines as closely as possible.
More implementation details are provided in Appendix~\ref{Appendixsec-B.3:Implementation Details}. All inference and training prompts are shown in Appendix~\ref{Appendixsec-B.2:Prompt}.

\section{Results and Discussions}
\subsection{Overall Performance}
\textbf{Local and web sources complement each other, making it valuable to explore both.}
In Table~\ref{tab:Overall Performance}, a comparison of the results of Vanilla and NoR on a series of LLMs shows that external knowledge improves answer quality.
In most cases, local sources alone perform better than web sources alone (cf. Appendix~\ref{Appendixsec-A.3: Different Retrieval Sources Strategies}), while using either source outperforms using no retrieval sources at all.
Furthermore, combining both local and web sources achieves better results than using either source individually, indicating that they provide complementary knowledge for answering questions.

\noindent\textbf{Simply concatenating knowledge from two sources fails to meet the external knowledge needs of LLMs.}
Analyzing the results on multi-hop QA, PrefRAG surpasses Vanilla${_{\textit{Mix}}}$, especially with a 25.6\% improvement on 2WikiMQA.
This reveals that PrefRAG enables a more thorough and effective utilization of both, rather than merely concatenating the two knowledge sources.
Moreover, on the simpler BioASQ-Y/N dataset, while the gap between our method and Vanilla${_{\textit{Mix}}}$ narrows, we still retain an advantage.
This is due to BioASQ-Y/N being relatively straightforward, typically requiring only a single-step inference to determine a \texttt{Yes/No} answer.

\noindent \textbf{PrefRAG outperforms SS-ARAG and MS-RAG through deeper, more effective and robust adaptive multi-source exploration.}
Compared to SS-ARAG, we observe that PrefRAG significantly surpasses SS-ARAG across all datasets, with improvements reaching up to 29.4\%. Even on the more challenging MusiQue dataset, PrefRAG still achieves a notable gain of up to 15.4\%. These results suggest that our method provides a more effective recipe for adaptive retrieval in a multi-source setting, rather than being limited to deep exploration within a single source.
Compared to MS-RAG, PrefRAG achieves significant improvements across all datasets, outperforming CRAG by up to 39.9\%, ReAct by up to 13.9\%, and ReAct${_{\textit{Mix}}}$ by up to 6.2\%.
We further analyze the underlying reasons behind these results.
\textit{Firstly}, CRAG's one-time retrieval approach lacks adaptive exploration capability.
\textit{Secondly}, ReAct is unable to foresee source characteristics because it relies on tool descriptions and parametric knowledge for source selection.
This leads to uncertain initial source selections and premature source abandonment once failed attempts, limiting thorough exploration. 
While ReAct${_{\textit{Mix}}}$ maximizes multi-source by concatenating both sources at each step, it introduces more noise that potentially impacts reasoning.
In contrast, PrefRAG examines local sources based on preset preferences and switches sources only after confirming knowledge quality, enhancing the robustness of retrieval selection.

\noindent \textbf{DPO effectively improves the ability of the model for preference-driven retrieval selection.}
By comparing the scores of GLM4-9B-chat and GLM4-9B-chat with DPO as end-to-end backbone models, we find that DPO significantly improves in-domain performance (+2.5\%) and out-of-domain performance (up to +3.1\%). This improvement trend remains consistent across both complex multi-hop and simple biomedical QA tasks.
This trend indicates that the trained model exhibits more competitive capabilities in selecting and switching retrieval sources, enabling more effective knowledge utilization for answer generation. Furthermore, the out-of-domain results demonstrate its strong generalization across diverse datasets.

\begin{table*}[ht]
\setlength{\tabcolsep}{3.5pt}
    \fontsize{8}{8.5}\selectfont
  \centering

    \begin{tabular}{ccccccccccccccc}
    \Xhline{1pt}
    \multirow{2}[2]{*}{\textbf{LLMs}} & \multirow{2}[2]{*}{\textbf{Methods}} & \multicolumn{4}{c}{\textbf{HotpotQA}} & \multicolumn{4}{c}{\textbf{2WikiMQA}} & \multicolumn{4}{c}{\textbf{MusiQue}} & \textbf{BioASQ-Y/N} \\
    \cmidrule(lr){3-6} \cmidrule(lr){7-10} \cmidrule(lr){11-14} \cmidrule(lr){15-15} &  & Acc.  & F1    & EM    & Avg.  & Acc.  & F1    & EM    & Avg.  & Acc.  & F1    & EM    & Avg.  & Acc. \\
    \hline
    \multirow{3}[0]{*}{Llama3.1-8B-Instruct} & \textbf{PrefRAG} &  \textbf{42.0} &  \textbf{51.1} & 38.8 & \multicolumn{1}{c}{ \textbf{44.0}} &  \textbf{42.0} & \textbf{43.2} &  \textbf{35.8} &  \textbf{40.3} &  \textbf{15.4} &  \textbf{21.0} &  \textbf{12.8} &  \textbf{16.4} & \textbf{89.6}  \\
    & \textit{w/o} Pref-AR & 41.0  & \underline{50.9}  &  \textbf{39.8} & \underline{43.9}  & 36.0  & 37.8  & 30.2  & 34.7  & \underline{13.6}  & 19.0  & 11.0  & 14.5  & \underline{81.4}  \\
    & \textit{w/o} Self-Reflection & \underline{41.6} & \underline{50.9} & \underline{39.6} & \textbf{44.0} & \underline{41.6} & \underline{42.1} & \underline{34.4} & \underline{39.4} & 13.2  & \underline{19.9} & \underline{12.2} & \underline{15.1} & \textbf{89.6} \\
    \hdashline
    \multirow{3}[0]{*}{GLM4-9B-chat-DPO} & \textbf{PrefRAG} &  \textbf{51.4} &  \textbf{57.0} &  \textbf{45.0} &  \textbf{51.1} &  \textbf{57.0} &  \textbf{56.0} &  \textbf{45.2} &  \textbf{52.7} &  \textbf{24.2} &  \textbf{30.0} &  \textbf{20.2} &  \textbf{24.8} &  \underline{89.6} \\
    & \textit{w/o} Pref-AR & 47.4  & 53.4  & 41.0  & 47.3  & 53.6  & 53.4  & 40.0  & 49.0  & 18.0  & 23.1  & 14.4  & 18.5  & 88.8  \\
    & \textit{w/o} Self-Reflection & \underline{49.4}  & \underline{56.0}  & \underline{42.6} & \underline{49.3}  & \underline{56.8}  & \underline{54.4}  & \underline{41.8}  & \underline{51.0}  & \underline{22.4}  & \underline{28.0}  & \underline{18.4}  & \underline{22.9}  & 
    \textbf{89.8}  \\
    \hdashline
    \multicolumn{1}{c}{\multirow{3}[0]{*}{GLM4-Plus}} & \multicolumn{1}{c}{\textbf{PrefRAG}} & \multicolumn{1}{c}{\textbf{59.0}} & \multicolumn{1}{c}{\textbf{68.4}} & \multicolumn{1}{c}{\textbf{55.0}} & \multicolumn{1}{c}{\textbf{60.8}} & \multicolumn{1}{c}{\textbf{79.6}} & \multicolumn{1}{c}{\textbf{76.7}} & \multicolumn{1}{c}{\textbf{65.2}} & \multicolumn{1}{c}{\textbf{73.8}} & \multicolumn{1}{c}{\textbf{32.2}} & \multicolumn{1}{c}{\textbf{39.4}} & \multicolumn{1}{c}{\textbf{27.4}} & \textbf{33.0}  & \textbf{94.0} \\
    & \textit{w/o} Pref-AR & 51.6  & 61.1  & 47.8 & 53.5  & 74.2  & 72.6  & 59.6  & 68.8  & 26.2  & 33.3  & 22.0  & 27.2  & 93.4  \\
    & \textit{w/o} Self-Reflection & \underline{57.6}  & \underline{67.3}  & \underline{53.8}  & \underline{59.6}  & \underline{78.6}  & \underline{74.8}  & \underline{62.8}  & \underline{72.1}  & \underline{32.0} & \underline{38.5}  & \underline{27.0} & \underline{32.5} & \underline{93.6}  \\
    \Xhline{1pt}
    \end{tabular}%
    \caption{{\bf Results (\%) of ablation study.} The "\textit{w/o} Pref-AR" means we omit the preference-driven retriever selection, and leave the LLM to choose a retrieval source by itself. The "\textit{w/o} Self-Reflection" means removing the answer assessment and directly using the first generated answer.}
    \label{tab:All ablation results}
\end{table*}%

\subsection{Ablation Study}
We conduct an ablation study on all datasets (cf. Appendix~\ref{Appendixsec-A.2: All Results of Ablation Study}) to analyze key components, with the main results shown in Table~\ref{tab:All ablation results}.
We observe that both "Pref-AR" and "Self-Reflection" play a crucial role, demonstrating the effectiveness of our preference-driven retrieval and self-reflection processes. In most cases, "Pref-AR" serves as the primary contributor, while self-reflection plays a secondary role.
The underlying reason for this phenomenon is that Pref-AR determines the quality of retrieved knowledge, directly impacting answer generation. Self-reflection's effectiveness is bounded by retrieval quality and model capabilities. Notably, when using larger models or DPO-trained models as the backbone, both components show increased effectiveness, with Pref-AR's primary role becoming more prominent. This improvement stems from enhanced model capabilities in question analysis, retrieval exploration, self-reflection, and instruction-following, strengthening the adaptive retrieval process.

\begin{figure}[htbp]
\includegraphics[width=\columnwidth]{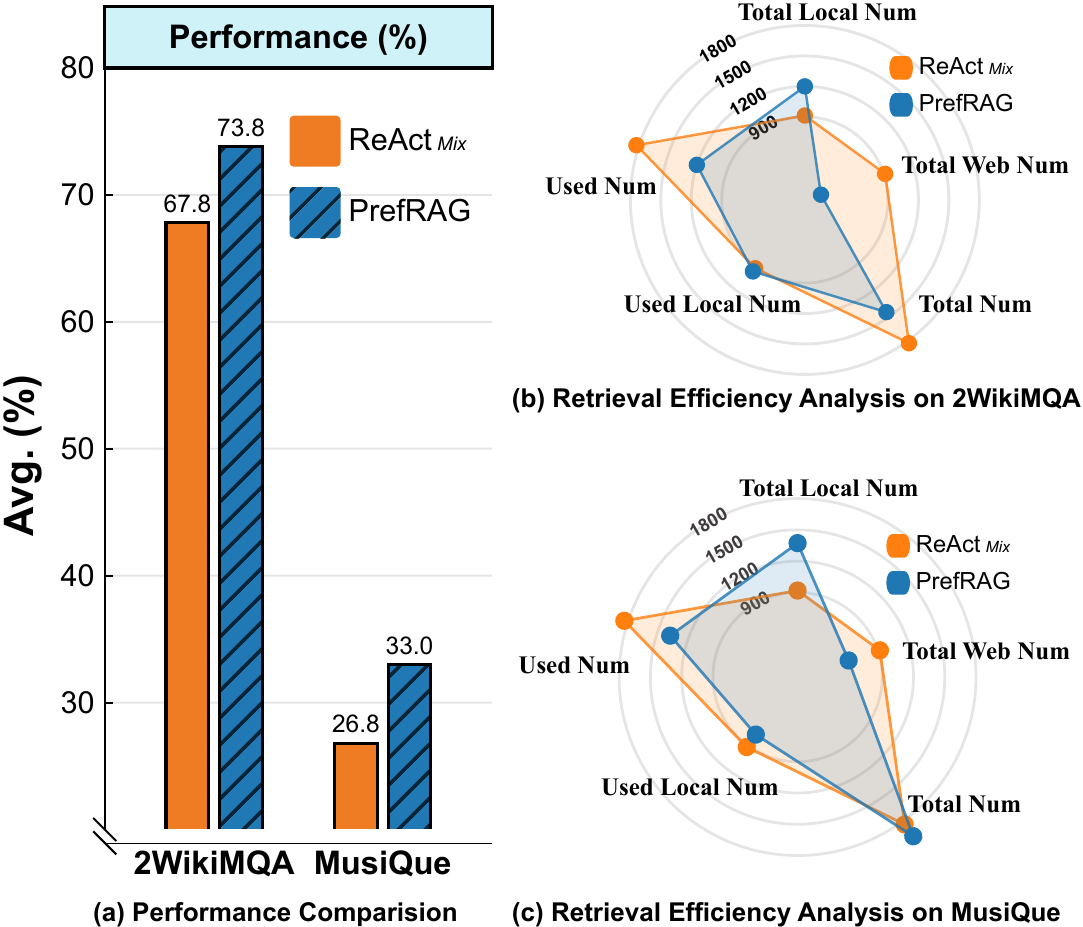}
\caption{\textbf{Retrieval count and performance analysis} on 2WikiMQA and MusiQue datasets.}
\label{fig:RC-P trade-off}
\end{figure}

\subsection{Efficiency and Performance Analysis}
An intuitive assumption is that directly concatenating all retrieved documents from multiple sources maximizes source perception.
However, our analysis demonstrates that PrefRAG offers significant advantages in both performance and retrieval efficiency compared to ReAct$_{\textit{Mix}}$ with a direct multi-source concatenation approach.
Fig.~\ref{fig:RC-P trade-off} shows that PrefRAG achieves superior performance through fewer total retrieval counts on 2WikiMQA and competitive retrieval counts with superior performance on MusiQue.
Notably, PrefRAG reasoning process requires significantly fewer retrieval counts ("Used Num") than ReAct$_{\textit{Mix}}$, indicating more precise source selection.
The reduced web retrieval counts demonstrate PrefRAG preference for local sources, making it particularly suitable for real-world applications requiring controlled knowledge retrieval (\S \ref{subsec:Real-World Applications of PrefRAG}).

\begin{figure}[htbp]
\includegraphics[width=\columnwidth]{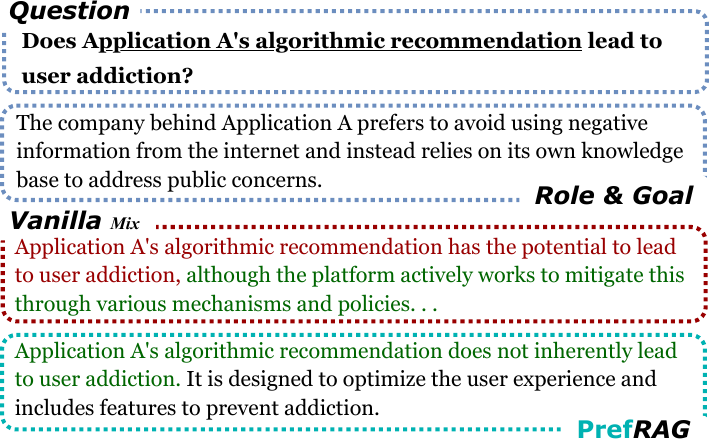}
\caption{\textbf{Examples of controllable knowledge retrieval.} "\textcolor{myred}{Red}" and "\textcolor{mygreen}{green}" represent desirable and undesirable information, respectively.}
\label{fig:Pref-example}
\end{figure}

\subsection{Real-World Applications of PrefRAG}
\label{subsec:Real-World Applications of PrefRAG}
\noindent \textbf{Controllable Knowledge Retrieval.} 
In real-world applications, AI systems accessing websites pose various risks~\citep{DBLP:journals/corr/abs-2310-19852}.
Some developers seek AI outputs aligned with their preferences, such as favorable product evaluations.
By providing controlled knowledge, PrefRAG can guide the system toward desired outputs for users.  Therefore, developing controllable knowledge retrieval RAG systems is essential for ensuring both accuracy and output preference control.
Specifically, PrefRAG enhances controllability by prioritizing local corpus retrieval before web access.
To demonstrate this, we create role-aligned scenarios using real-world information.
Sensitive information has been anonymized.
As Figure~\ref{fig:Pref-example} shows, PrefRAG prioritizes retrieval from its controlled knowledge corpus with intended promotional materials, while avoiding potentially unfavorable external content.
It only accesses the web when the corpus lacks relevant information (cf. Appendix~\ref{Appendixsec-D: Controllable Knowledge Retrieval}).
In contrast, providing both sources directly (i.e., Vanilla$_{\textit{mix}}$) may generate undesirable content.

\section{Conclusion}
In this work, we identify the limitations of ARAG systems in effectively and controllably exploring diverse sources. We introduce PrefRAG, a MS-ARAG framework that enables in-depth and controllable adaptive exploration of different retrieval sources through preference-driven adaptive retrieval and self-reflection. We conduct multi-dimensional studies to confirm the superiority of PrefRAG and present its controllable knowledge retrieval ability in realistic scenarios.

\section{Limitations}
Extensive empirical studies have demonstrated that PrefRAG exhibits high performance, retrieval efficiency, and great potential for controllable knowledge retrieval in real-world applications. Nevertheless, certain limitations remain that deserve further attention. Addressing these limitations will be a key focus in future work.

\noindent \textbf{Challenges in Fine-Grained Retrieval Sources and Multiple Preferences Integration.}
In this work, we explored system performance using two widely used retrieval sources: local and web. However, we did not analyze PrefRAG's performance under more fine-grained retrieval source configurations and more preset preferences.
For example, the local retrieval source could be further subdivided into sources from more specialized domains, and web sources could be divided based on different types of search engines.
Our system theoretically supports integration with more retrieval sources and can switch between them based on our selection strategy when making retrieval decisions. However, incorporating multiple preset preferred sources could lead to preference conflicts, posing significant challenges.
Moving forward, we anticipate developing an interaction strategy for multiple retrieval sources and diverse preference requirements.
This could be an effective approach to aligning PrefRAG with the more complex preference-driven retrieval requirements in real-world applications.

\noindent\textbf{Foundational Model Dependency.}
Smaller-size models, limited by the size of their parameter knowledge, suffer from reduced reasoning ability. This inherent limitation can lead to low-quality retrieval queries. However, the quality of our retrieval source selection depends on the quality of the retrieval queries generated by the model. Although we place the retrieved documents within the context and feed them back to the model as feedback, this does not fully eliminate the impact of the model's inherent capability limitations. Therefore, further research into enhancing the ability of smaller-size models to generate high-quality queries will further improve the performance of the PrefRAG system.

\bibliography{acl}

\clearpage

\appendix
\section*{Appendix}
\label{sec:appendix}

\startcontents[sections]
\printcontents[sections]{l}{1}{\setcounter{tocdepth}{2}}

\section{Additional Experimental Results}
\label{Appendixsec-A: Additional Experimental Results}

\subsection{More Results of Overall Performance}
\label{Appendixsec-A.1: More Results of Overall Performance}
Table~\ref{tab-app:Overall Performance} presents more results of overall performance.
Compared to Table~\ref{tab:Overall Performance}, we supplement the results of Vanilla RAG and LLM without retrieval based on the Llama3.1-70B-Instruct model.
Here, Vanilla RAG includes using only local retrieval sources and using both local and web retrieval sources.
Furthermore, we provide results for ReAct \textit{w/} LR $\&$ WR on all models.
The trends and conclusions of these results are similar to those in Table~\ref{tab:Overall Performance}.
These results further demonstrate the significant superiority, effectiveness, and robustness of PrefRAG.

\subsection{All Results of Ablation Study}
\label{Appendixsec-A.2: All Results of Ablation Study}
In Table~\ref{tab-app:All ablation results}, we present the results of the ablation study on all models. We observe that preference-driven retrieval serves as the primary contributor, while self-reflection plays a secondary role. This aligns with the conclusions and trends in Table~\ref{tab:All ablation results}. We note that some smaller-size parameter models struggle to effectively perform retrieval source selection due to insufficient instruction-following capabilities.
Through DPO training, smaller-size parameter models can select retrieval sources more accurately and robustly, thereby consistently gathering more effective context.
This higher-quality context further enhances the ability of smaller-size parameter models to execute more effective self-reflection processes.
These results and trends confirm the effectiveness of the preference-driven retrieval and the self-reflection process of PrefRAG, as well as the effectiveness of our automated training data construction pipeline and training strategy.
\subsection{Different Retrieval Sources Strategies}
\label{Appendixsec-A.3: Different Retrieval Sources Strategies}
We conduct a pilot experiment to analyze the impact of using multiple types of retrieval sources on the performance of RAG systems. In our work, we utilize two of the most mainstream retrieval sources with distinct content characteristics: local and web retrieval sources.

As shown in Table~\ref{tab:All results of Vanilla with different retrieval sources}, in most cases, the carefully curated local retrieval source provides greater performance improvements for RAG systems compared to the open and real-time web retrieval source. Furthermore, simply concatenating documents retrieved from both sources can yield higher performance than using either source alone. This indicates that the knowledge from the two types of retrieval sources can complement each other.
Investigating effective and appropriate methods to harness knowledge from multiple sources represents a valuable research direction.

\subsection{Different PrefRAG Strategies}
\label{Appendixsec-A.4: Different PrefRAG Strategies}
\begin{figure}[htbp]
\includegraphics[width=\columnwidth]{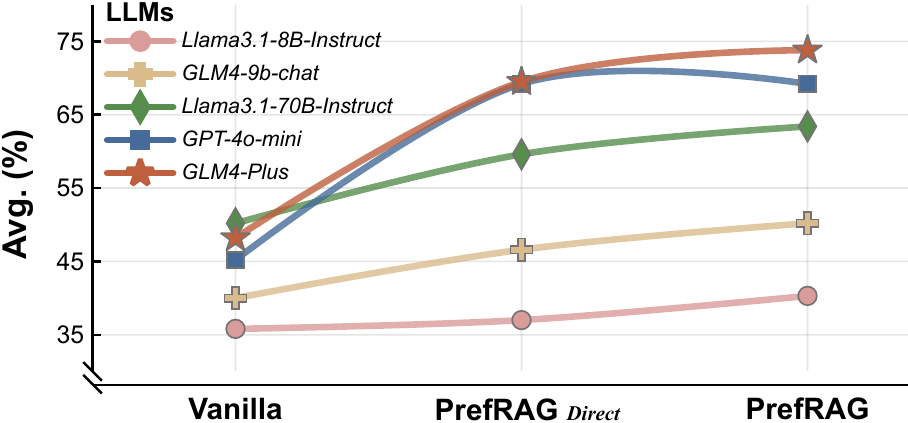}
\caption{{\bf Different Strategies of PrefRAG on 2WikiMQA.} "Vanilla" represents "Vanilla${_{\textit{Mix}}}$ \textit{w/} LR $\oplus $ WR".}
\label{fig:Different Strategies}
\end{figure}
\noindent We investigate the impact of different preference strategies. Specifically, we test a direct approach that explicitly states preferred retrieval sources in multi-source tool descriptions.
We compare PrefRAG with PrefRAG$_{\textit{Direct}}$, which employs a simplified preference-driven strategy.
Unlike PrefRAG, PrefRAG$_{\textit{Direct}}$ integrates preference-driven adaptive retrieval into the overall prompt as a linguistic description.
As shown in Figure~\ref{fig:Different Strategies}, PrefRAG$_{\textit{Direct}}$ achieves notable performance gains over the preference-free multi-source retrieval baseline, Vanilla RAG, especially when the backbone model is strong.
This implies that incorporating preferences, regardless of the strategy, facilitates a more structured exploration of multiple sources.
However, PrefRAG$_{\textit{Direct}}$ still falls short of PrefRAG, as it incorporates $D_{k,t}$ into the $o_{t}$ within the current iteration $\tau_{t}$ but can only adjust the retrieval source in the next iteration $\tau_{t+1}$. In other words, PrefRAG enables more timely corrections of inappropriate retrieval sources.

\subsection{Different top-\texorpdfstring{$k$}{k} Values and Retrievers}
\label{Appendixsec-A.5: Different top-k Values and Retrievers}
The choice of top-$k$ in RAG systems controls the number of documents fed into the LLM, thereby influencing the quality of the final answer.
To validate that our method achieves significant performance improvements across various top-$k$ values, we conduct experiments with multiple top-$k$ settings.

In Table~\ref{tab-app:Top-k}, we observe that PrefRAG maintains a notable performance advantage across different top-$k$ values, particularly on complex multi-hop questions.
On the BioASQ-Y/N dataset, which requires only simple reasoning, we find that an appropriate top-$k$ can elicit optimal performance, while high top-$k$ values may introduce noise, thereby degrading the final answer quality.
Additionally, we find that a larger top-$k$ yields higher performance on the more challenging dataset (e.g., MusiQue). For relatively simpler datasets like HotpotQA and BioASQ-Y/N, we recommend researchers use a moderate top-$k$.

The choice of different retrievers also affects the quality of documents fed into the LLM, affecting the final answer quality.
Therefore, we conduct experiments on two mainstream retrieval approaches, i.e., sparse retrieval and dense retrieval, to demonstrate the robustness and generalizability of PrefRAG. As shown in Table~\ref{tab-app:Retriever}, we find that PrefRAG achieves comparable performance with different types of retrievers.
This phenomenon suggests that PrefRAG is compatible with various retrievers, demonstrating its robustness.

\subsection{Retrieval Counts Details}
\label{Appendixsec-A.6: Retrieval Counts Details}
All results of the efficiency and performance analysis are presented in Table~\ref{tab-app:Total counts of retrieval on HotpotQA}, Table~\ref{tab-app:Efficiency and accuracy trade-off on HotpotQA}, Table~\ref{tab-app:Total counts of retrieval on 2WikiMQA}, Table~\ref{tab-app:Efficiency and accuracy trade-off on 2WikiMQA}, Table~\ref{tab-app:Total counts of retrieval on MusiQue}, Table~\ref{tab-app:Efficiency and accuracy trade-off on MusiQue}, Table~\ref{tab-app:Total counts of retrieval on BioASQ-Y/N}, and Table~\ref{tab-app:Efficiency and accuracy trade-off on BioASQ-Y/N}.
The specific values in Figure~\ref{fig:RC-P trade-off} are presented in Tables~\ref{tab-app:Total counts of retrieval on 2WikiMQA}, Table~\ref{tab-app:Efficiency and accuracy trade-off on 2WikiMQA}, Table~\ref{tab-app:Total counts of retrieval on MusiQue}, and Table~\ref{tab-app:Efficiency and accuracy trade-off on MusiQue}.

Specifically, "\textbf{Total Local Num}" represents the total number of local retrieval counts, while "\textbf{Total Web Num}" denotes the total number of web retrieval counts.
Notably, "\textbf{Total Web Num}" also represents the number of web retrieval counts used for inference.
"\textbf{Total Num}" refers to the overall number of retrievals, which is the sum of "\textbf{Total Local Num}" and "\textbf{Total Web Num}".
"\textbf{Used Local Num}" indicates the number of local retrievals used for inference. Since local retrieval requires assessing whether the retrieved knowledge is useful and contributes to knowledge augmentation, some iterations may switch to web retrieval. When switching to the web source, passages retrieved from the local source are no longer included in the context for inference.
"\textbf{Used Num}" represents the total number of retrievals used for inference.

We compare PrefRAG and ReAct \textit{w/} LR $\oplus$ WR in five dimensions and performance aspects. Notably, this is not an entirely fair comparison, as ReAct \textit{w/} LR $\oplus$ WR incorporates both local and web retrieval results into the context at each iteration, whereas PrefRAG must choose between the two sources and include only one in the context for inference. Despite this, PrefRAG consistently outperforms ReAct \textit{w/} LR $\oplus$ WR in overall performance in most cases.
This trend suggests that preference-driven retrieval, which carefully selects the most effective retrieval source, is superior to indiscriminately incorporating multiple sources in every iteration.

Alongside its performance advantage, we also observe a significant reduction in both "Total Num" and "Used Num", indicating that PrefRAG reduces unnecessary retrieval attempts and retrievals included in the context, thereby improving retrieval efficiency.
Furthermore, PrefRAG demonstrates the ability to conduct a deeper exploration of the preferred retrieval source when appropriate. In some cases, its "Total Local Num" surpasses that of ReAct \textit{w/} LR $\oplus$ WR. However, its "Used Local Num" decreases significantly, and this reduction exceeds the increase in "Total Local Num", suggesting that PrefRAG not only explores more thoroughly but also precisely identifies relevant retrievals for inference, minimizing noise and token overhead from ineffective documents.
More importantly, PrefRAG significantly reduces "Total Web Num" through preference-driven retrieval, effectively lowering the risk of exposing RAG systems to undesirable web content in controlled retrieval settings, something ReAct \textit{ w /} LR $\oplus$ WR fails to achieve.

\section{More Experimental Setup Details}
\label{More Details of Experimental Setup}
We summarize dataset statistics and all the experimental settings in Table~\ref{tab-app:Dataset statistics and settings}.
\subsection{Datasets}
\label{Appendixsec-B.1:Datasets}
For multi-hop QA datasets, we use the test sets released by
~\citep{DBLP:conf/acl/TrivediBKS23}
, each dataset containing 500 randomly selected QA pair samples. Additionally, for the BioASQ~(\citealp{DBLP:journals/bmcbi/TsatsaronisBMPZ15}; \citealp{krithara2023bioasq};
\citealp{DBLP:conf/acl/Xiong0LZ24}) dataset, we select the \texttt{Yes/No} questions in the ground truth test set of Task B from the most recent five years (2019-2023), including 500 questions in total.

\subsection{Prompts}
\label{Appendixsec-B.2:Prompt}
All PrefRAG prompts are presented in Table~\ref{tab-app:Overall prompt}, Table~\ref{tab:Input Variables in the Overall Prompt},
Table~\ref{tab-app:Preference-driven Retrieval Source Selection prompt},
Table~\ref{tab-app:Input Variables in the Preference-driven Retrieval Source Selection Prompt},
Table~\ref{tab-app:Get labels for preferred retrieval}
,
and Table~\ref{tab-app:Input Variables in prompt for obtaining preferred retrieval labels in the training data}.
In Table~\ref{tab-app:Overall prompt}, we provide the overall prompt, which includes the adaptive retrieval process (excluding the preference-driven retrieval source selection stage) and the self-reflection process.
Table~\ref{tab:Input Variables in the Overall Prompt} explains all input variables in the overall prompt.
Table~\ref{tab-app:Preference-driven Retrieval Source Selection prompt} presents the preference-driven retrieval source selection prompt, denoted as $\text{Instruct}_{\text{Sel}}$ in Equation (\ref{eq4.2.3:Selection}), and Table~\ref{tab-app:Input Variables in the Preference-driven Retrieval Source Selection Prompt} explains its input variables.

For prompts used during training, Table~\ref{tab-app:Get labels for preferred retrieval} provides the prompt for obtaining preferred retrieval labels, with Table~\ref{tab-app:Input Variables in prompt for obtaining preferred retrieval labels in the training data} detailing its input variables.

\subsection{Implementation Details}
\label{Appendixsec-B.3:Implementation Details}
For prompts, we consider that answer format variations may impact evaluation results. To ensure a fair comparison, we align the answer format in the prompts for generating responses across all baselines as closely as possible.

In our experiment, we encourage the system to minimize costs while achieving better results.
Therefore, we set the maximum number of iterations for the adaptive retrieval process to 3.
During the self-reflection process, we limit the maximum number of supplementary retrievals and entries into the preference-driven adaptive retrieval process to one.
This means that the system will directly generate an answer when the self-reflection token is labeled as non-"CORRECT" for the second time.
Additionally, we observe that agent-based frameworks (i.e., ReAct) might, in extreme cases, fail to provide a final answer even after reaching the maximum iterations.
Notably, the response format of agent-based methods is inherently uncertain.
In a few cases, they may fail to produce an answer, they may fail to produce an answer.
To address this, we employ a forced answer generation mechanism: if no answer is provided in the final iteration, the system is instructed to generate an answer based on the existing context.

We implement Self-RAG and CRAG using LangChain\footnotemark[4] framework. For FLARE and ReAct, we follow their official code implementations. All implementations utilize the same local corpus and retriever as our method for fair comparison. For CRAG and ReAct, we configure DuckDuckGo as the web source, maintaining consistency with PrefRAG.
\footnotetext[4]{\url{https://github.com/langchain-ai/langgraph}}

\section{DPO Data Construction Details}
\label{Appendixsec-C: DPO Data Construction Details}

We randomly sample 15,000 instances from the training set of the 2WikiMQA dataset to construct the training data.
First, we use GLM-9B-chat to perform the adaptive retrieval process starting from the $q$.
During the iteration $\tau=\tau_{2},\dots,\tau_{n}$, we configure nine different combinations of model hyper-parameter by varying the temperature and top-p values across three different settings $\left\{0.1,0.5,0.9\right\}$, ensuring a clearer distinction between positive and negative samples.
These combinations generate nine predictions during the retrieval source selection process. Each prediction includes a CoT analysis and a status value (i.e., \texttt{True} or \texttt{False}), indicating whether to switch retrieval source.
Note that since a single sample generates these predictions across multiple iterations, we also perform random sampling to ensure that the final training samples contain no duplicates and cover data from various iterations.
Concurrently, we also use a larger-size parameter model, GLM4-Plus, of the same series to output a gold label for retrieval source selection.
In detail, we present the prompt for generating predictions in Table~\ref{tab-app:Preference-driven Retrieval Source Selection prompt} and its input variables in Table~\ref{tab-app:Input Variables in the Preference-driven Retrieval Source Selection Prompt}. The input variables of the prompt together constitute the input $x$ in the training data.
Next, we compare the nine predictions generated by GLM4-9B-chat with the gold label.
Instances with matching status values form the positive candidate set, while those with differing values form the negative candidate set.

We then use the prompt in Table~\ref{tab-app:Get labels for preferred retrieval} to compare the data in the positive candidate set with the gold label and employ GLM4-Plus to select the best instance as the positive sample for training. For the negative sample, we randomly select one instance from the negative candidate set.
Additionally, we notice that in the 2WikiMQA dataset, over approximately 70\% labels generated by GLM4-Plus have a status value of \texttt{True}. To simulate the real distribution, we select 3,000 instances with a \texttt{True} status value as positive samples $y^{+}$ and 1,000 instances with a \texttt{False} status value as negative samples $y^{-}$, resulting in 4,000 training samples, denoted as $\left \{ x,y^{+},y^{-} \right \}\sim \mathcal{D} $.

\section{Controllable Knowledge Retrieval}
\label{Appendixsec-D: Controllable Knowledge Retrieval}
We construct two types of controllable retrieval scenarios.
Specifically, we collect real-world questions and conduct searches on the open web. The retrieved positive answers are compiled into our corpus. To simulate a more realistic retrieval process, we merge this corpus with the 2WikiMQA corpus to form the final retrieval corpus.

Table~\ref{tab-app: PrefRAG Example for Pref1} presents a controllable response example where the user expects the answer to be generated using knowledge from the local retrieval source.
In these cases, specific roles expect the RAG system to rely on knowledge from more controllable local retrieval sources for the final answer while avoiding unfavorable information from the web.
Table~\ref{tab-app: PrefRAG Example for Pref2} presents examples where web sources supplement knowledge.
Here, specific roles expect the RAG system to supplement local retrieval when its knowledge is insufficient by leveraging web sources.
These examples demonstrate that PrefRAG enables users with controllable response needs to prioritize retrieving knowledge from local sources, such as carefully curated brand information.
At the same time, it can flexibly incorporate web knowledge when local sources are insufficient.
This capability allows RAG to expand its retrieval scope while maintaining control over the retrieval process, thereby improving answer quality and mitigating risks associated with unreliable web information.
Consequently, PrefRAG enhances both the adaptability and reliability of RAG systems in real-world applications.

\section{More Retrieval Sources}
\label{Appendixsec-E: More Retrieval Sources}
In our work, we primarily conduct experiments using two classic retrieval sources with distinct characteristics.
However, PrefRAG can support multiple retrieval sources (more than two) along with one predefined retrieval preference in practical applications.
For example, PrefRAG can integrate four retrieval sources, $S_{1}$, $S_{2}$, $S_{3}$, and $S_{4}$, with one designated as the preferred retrieval source, such as $S_{1}$.
This requires adjustments to the operations in the two stages of the preference-driven retrieval decision process.

Specifically, in the Retrieve-or-Generate stage, the action space is no longer limited to a single "\texttt{Search\_Engine}" action but instead includes four actions: "\texttt{Search\_S1}", "\texttt{Search\_S2}", "\texttt{Search\_S3}", and "\texttt{Search\_S4}".
The model needs to determine whether to continue retrieval and which source to retrieve based on the existing context.
For example, the model determines to continue the retrieval and select "\texttt{Search\_S2}" at this stage.
We retrieve $S_{1}$ following the predefined retrieval preference.
If $\text{Instruct}_{\text{Sel}}$ determines that a source switch is necessary, we then perform retrieval using "$S_{2}$".

\section{Case Study}
\label{Appendixsec-F: Case Study}

We conduct a case study, and QA examples of PrefRAG are presented in Table~\ref{tab-app:PreRAG examples on HotpotQA dataset}
,~\ref{tab-app:ReAct examples on Hotpot dataset (Local or Web)}
, and~\ref{tab-app:PreRAG examples on  2WikiMQA dataset}. 

In Table~\ref{tab-app:PreRAG examples on HotpotQA dataset}, given the original query $q$, "\textit{In what year did the Danish plant ecologist who assisted a Danish chemist, famous for the introduction of the concept of pH, die?}", PrefRAG first analyzes the $q$ and formulates a reasoning thought: "\textit{I need to identify the Danish plant ecologist who assisted a Danish chemist famous for introducing the concept of pH}".
In iteration $\tau_{1}$, PrefRAG retrieves information about the Danish chemist who introduced the concept of pH and identifies him as Søren Peder Lauritz Sørensen.
In iteration $\tau_{2}$, PrefRAG refines its thought: "\textit{Now I need to find the Danish plant ecologist who assisted him}".
To enhance retrieval accuracy, PrefRAG incorporates the chemist’s name into a new subquery: "\textit{Danish plant ecologist who assisted Søren Peder Lauritz Sørensen}".
However, in the next iteration, PrefRAG considers that the retrieval "\textit{did not provide specific information about a Danish plant ecologist who assisted Søren Peder Lauritz Sørensen}".
It then strategizes its goal for the next iteration: "\textit{I need to consider if there might be a misunderstanding in the question or if the information is not readily available}".
At this iteration, the system attempts to generate an answer $\alpha$, accompanied by a self-reflection label {\icorr}, explanation, and improvement suggestions. The self-reflection label correctly identifies that "Not available" is an incorrect answer.
In the explanation and improvement suggestions, the system reflects on the error, noting that the lack of available information on who assisted Søren Peder Lauritz Sørensen prevented it from determining the year of death. It also suggests further historical research or seeking expert consultation in Danish scientific history.
A supplementary retrieval is then conducted, which reveals that Carsten Erik Olsen assisted Søren Peder Lauritz Sørensen and provides his birth and death years.
With this newly acquired knowledge, the model successfully identifies Carsten Erik Olsen as the Danish plant ecologist in the original query $q$.
In the \textbf{Final Answer}, PrefRAG correctly states that Carsten Erik Olsen passed away in \textbf{1974} and assigns the self-reflection label as {\corr}. The improvement suggestion is: "\textit{None needed, the answer is accurate based on the information found}".

For comparison, Table~\ref{tab-app:ReAct examples on Hotpot dataset (Local or Web)} presents how ReAct approaches the same question.
In some cases, ReAct initially retrieves information from web sources, causing it to miss valuable knowledge from carefully curated local sources.
In iteration $\tau_{1}$, ReAct correctly identifies that the Danish chemist famous for introducing the concept of pH is Søren Sørensen.
However, in iteration $\tau_{2}$, it retrieves information from the web suggesting that Thorvald (Thorwald) Julius Sørensen might be connected to the Danish chemist, which is incorrect. Due to this misidentification, ReAct ultimately provides an incorrect year of death for the Danish plant ecologist.
By comparing PrefRAG and ReAct, we find that ReAct’s initial choice of retrieval sources exhibits a degree of uncertainty.
In contrast, PrefRAG follows a preset preference as a guide.
Additionally, PrefRAG leverages self-reflection to critically assess its answers, refine subsequent retrieval and reasoning, and generate more reliable and high-quality responses.

Table~\ref{tab-app:PreRAG examples on  2WikiMQA dataset} also presents cases where PrefRAG provided the correct answer on the first attempt.
Given the original query $q$, "\textit{Which one was established first, Grouplogic or Inbios?}", we observe that PrefRAG follows a clear problem-solving approach: "\textit{I need to find the years of establishment of Grouplogic and Inbios to determine which one was established first}".
It then retrieves "\textit{GroupLogic, Inc., founded in 1988}" in iteration $\tau_{1}$ and "\textit{InBios International, Inc. was founded in 1996}" in iteration $\tau_{2}$.
Ultimately, PrefRAG correctly identifies \textbf{Grouplogic} as the answer, with a self-reflection label of \corr.

\clearpage

\begin{table*}[ht!]
\setlength{\tabcolsep}{3pt}
    \fontsize{9}{10}\selectfont
  \centering
    \begin{tabular}{ccccc}
    \toprule
    \multicolumn{1}{c}{\textbf{Settings}} & \multicolumn{1}{c}{\textbf{HotpotQA}} & \multicolumn{1}{c}{\textbf{2WikiMQA}} & \multicolumn{1}{c}{\textbf{MusiQue}} & \multicolumn{1}{c}{\textbf{BioASQ-Y/N}} \\
    \midrule
    \multicolumn{5}{c}{\textit{Dataset statistics}} \\
    \multicolumn{1}{c}{\#\,\,Samples used for evaluation} & \multicolumn{1}{c}{500} & \multicolumn{1}{c}{500} & \multicolumn{1}{c}{500} & \multicolumn{1}{c}{500} \\
    \midrule
    \multicolumn{5}{c}
    {\textit{Evaluation settings}} \\
    \multicolumn{1}{c}{Metric} & \multicolumn{1}{c}{Accuracy, F1, EM} & \multicolumn{1}{c}{Accuracy, F1, EM} & \multicolumn{1}{c}{Accuracy, F1, EM} & \multicolumn{1}{c}{Accuracy} \\
    \midrule
    \multicolumn{5}{c}{\textit{Retrieval settings}} \\
    \multicolumn{1}{c}{Corpus} & \multicolumn{1}{c}{~\citep{DBLP:conf/acl/TrivediBKS23}} & \multicolumn{1}{c}{~\citep{DBLP:conf/acl/TrivediBKS23}} & \multicolumn{1}{c}{~\citep{DBLP:conf/acl/TrivediBKS23}} & \multicolumn{1}{c}{PubMed} \\
    \#\,\,Documents in Corpus & 5233329 & 139416 & 430139 & 23898701 \\
    Retriever & BM25, Dense & BM25, Dense & BM25, Dense & \multicolumn{1}{c}{BM25, Dense} \\
    top-$k$ & 3,5,7 & 3,5,7 & 3,5,7 & 3,5,7 \\
    \midrule
    \multicolumn{5}{c}
    {\textit{LLM settings}} \\
    \#\,\,Types of LLMs & 5 & 5 & 5 & 5 \\
    \bottomrule
    \end{tabular}%
    \caption{{\bf Dataset statistics and experimental settings of different datasets.}}
  \label{tab-app:Dataset statistics and settings}%
\end{table*}%

\begin{table*}[ht!]
\renewcommand{\arraystretch}{0.9}
\setlength{\tabcolsep}{3pt}
    \fontsize{7}{8}\selectfont
  \centering
  
    \begin{tabular}{lccccccccccccc}
    \toprule
          \multirow{2}[2]{*}{\textbf{Methods \& LLMs}} & \multicolumn{4}{c}{\bf HotpotQA} & \multicolumn{4}{c}{\bf 2WikiMQA} & \multicolumn{4}{c}{\bf MuSiQue} & {\bf BioASQ-Y/N}\\
\cmidrule(lr){2-5}\cmidrule(lr){6-9}\cmidrule(lr){10-13}\cmidrule(lr){14-14}        & Acc. & F1 & EM & Avg. & Acc. & F1 & EM & Avg. & Acc. & F1 & EM & Avg. & Acc. \\
    \hline
    \rowcolor[rgb]{ .851,  .851,  .851} \multicolumn{14}{c}{\textbf{\# Baselines without Retrieval (NoR) \#}} \\
    \multicolumn{14}{c}{\textit{Open-source LLMs}} \\
    Llama3.1-8B-Instruct & 22.6 & 28.7 & 23.0 & 24.8 & 27.4 & 30.7 & 26.4 & 28.2 & 3.6 & 9.4 & 3.2 & 5.4 & 77.8 \\
GLM4-9B-chat & 18.4 & 23.5 & 17.4 & 19.8 & 25.6 & 29.6 & 25.0 & 26.7 & 3.0 & 8.8 & 2.6 & 4.8 & 74.0 \\
Llama3.1-70B-Instruct & 32.4 & 41.5 & 31.4 & 35.1 & 33.8 & 37.9 & 32.6 & 34.8 & 8.0 & 14.6 & 7.4 & 10.0 & 87.0 \\
\hdashline
    \multicolumn{14}{c}{\textit{Proprietary LLMs}} \\
    GPT-4o-mini & 29.8 & 38.4 & 28.6 & 32.3 & 29.2 & 32.6 & 26.6 & 29.5 & 7.6 & 15.4 & 5.0 & 9.3 & 86.6 \\
    GLM4-Plus & 30.2 & 38.3 & 29.8 & 32.8 & 30.4 & 35.2 & 29.6 & 31.7 & 8.2 & 15.8 & 7.2 & 10.4 & 81.8  \\
    \hline
    \rowcolor[rgb]{ .851,  .851,  .851} \multicolumn{14}{c}{\textbf{\# Vanilla RAG (Vanilla) \#}} \\
    \multicolumn{14}{c}{\textit{Only local retrieval source (Vanilla \textit{w/ LR})}} \\
    Llama3.1-8B-Instruct  & 36.4 & 45.6 & 34.4 & 38.8 & 31.2 & 35.4 & 30.2 & 32.3 & 6.4 & 12.2 & 5.6 & 8.1 & 85.8 \\
    GLM4-9B-chat  & 34.8 & 44.4 & 34.2 & 37.8 & 34.4 & 38.8 & 33.8 & 35.7 & 8.2 & 15.0 & 7.0 & 10.1 & 87.2 \\
    Llama3.1-70B-Instruct & 42.6 & 53.4 & 42.6 & 46.2 & 45.2 & 48.2 & 43.0 & 45.5 & 11.4 & 18.4 & 10.6 & 13.5 & 89.4 \\
    GPT-4o-mini  & 45.0 & 53.8 & 41.2 & 46.7 & 40.2 & 44.2 & 38.6 & 41.0 & 11.2 & 19.2 & 8.8 & 13.1 & 89.6 \\
    GLM4-Plus  & 46.4 & 56.7 & 45.8 & 49.6 & 45.6 & 48.9 & 43.0 & 45.8 & 15.4 & 23.5 & 13.8 & 17.6 & 89.8 \\
    \hdashline
    \multicolumn{14}{c}{\textit{Concatenating both local and web retrieval source (Vanilla${_{\textit{Mix}}}$ w/ LR $\oplus $ WR)}} \\
    Llama3.1-8B-Instruct  & 41.6 & 53.9 & 41.2 & 45.6 & 35.4 & 39.3 & 32.6 & 35.8 & 9.0 & 16.0 & 8.0 & 11.0 & 89.6 \\
    GLM4-9B-chat  & 40.8 & 51.3 & 39.0 & 43.7 & 38.8 & 43.7 & 37.4 & 40.0 & 9.0 & 16.7 & 8.4 & 11.4 & 91.0 \\
    Llama3.1-70B-Instruct  & 47.2 & 59.9 & 46.8 & 51.3 & 49.6 & 54.0 & 47.0 & 50.2 & 13.4 & 21.4 & 12.6 & 15.8 & 93.2 \\
    GPT-4o-mini  & 47.4 & 58.0 & 44.6 & 50.0 & 45.8 & 49.1 & 40.6 & 45.2 & 13.2 & 21.3 & 11.4 & 15.3 & 92.2 \\
    GLM4-Plus  & 49.6 & 61.1 & 48.4 & 53.0 & 48.4 & 51.7 & 44.6 & 48.2 & 13.6 & 23.9 & 13.2 & 16.9 & 93.6 \\
    \hline
    \rowcolor[rgb]{ .851,  .851,  .851} \multicolumn{14}{c}{\textbf{\# Single-Source ARAG (SS-ARAG) \#}} \\
    FLARE $_{\text{GLM4-Plus}}$ & 46.4 & 51.8 & 41.8 & 46.7 & 49.4 & 45.9 & 37.8 & 44.4 & 16.6 & 21.9 & 14.4 & 17.6 & 77.2 \\
    Self-RAG $_{\text{GLM4-Plus}}$ & 45.0 & 54.5 & 43.6 & 47.7 & 32.4 & 36.7 & 30.2 & 33.1 & 15.4 & 24.3 & 13.2 & 17.6 & 82.8 \\
    \hline
    \rowcolor[rgb]{ .851,  .851,  .851} \multicolumn{14}{c}{\textbf{\# Multi-Source RAG (MS-RAG) \#}} \\
CRAG $_{\text{GLM4-Plus}}$ & 41.8 & 50.1 & 37.8 & 43.2 & 35.2 & 37.6 & 29.0 & 33.9 & 11.6 & 17.4 & 8.8 & 12.6 & 89.0 \\
ReAct \textit{w/ LR $\&$ WR} $_{\text{Llama3.1-8B-Instruct}}$ & 39.4 & 50.0 & 37.6 & 42.3 & 38.8 & 39.7 & 32.0 & 36.8 & 13.8 & 18.4 & 9.6 & 13.9 & 87.2 \\
ReAct \textit{w/ LR $\&$ WR} $_{\text{GLM4-9B-chat}}$ & 44.8 & 54.1 & 40.2 & 46.4 & 51.6 & 51.1 & 38.8 & 47.2 & 16.0 & 22.1 & 12.6 & 16.9 & 89.2 \\
ReAct \textit{w/ LR $\&$ WR} $_{\text{Llama3.1-70B-Instruct}}$ & 50.2 & 60.7 & 48.8 & 53.2 & 69.4 & 68.4 & 60.4 & 66.1 & 26.6 & 33.5 & 25.0 & 28.4 & 93.8 \\
ReAct \textit{w/ LR $\&$ WR} $_{\text{GPT-4o-mini}}$ & 51.8 & 60.3 & 47.0 & 53.0 & 72.2 & 69.9 & 55.6 & 65.9 & 19.0 & 25.6 & 14.6 & 19.7 & 91.0 \\
ReAct \textit{w/ LR $\&$ WR} $_{\text{GLM4-Plus}}$ & 50.0 & 59.7 & 46.2 & 52.0 & 64.2 & 63.8 & 51.8 & 59.9 & 23.2 & 30.6 & 18.4 & 24.1 & 91.8 \\
ReAct${_{\textit{Mix}}}$ \textit{w/ LR $\oplus$ WR} $_{\text{GLM4-Plus}}$ & 56.6 & 67.0 & 53.6 & 59.1 & 73.8 & 70.5 & 59.0 & 67.8 & 25.8 & 33.3 & 21.2 & 26.8 & 93.2 \\
    \hline
    \rowcolor[rgb]{0.85, 0.985, 0.985} \multicolumn{14}{c}{\textbf{\# Ours \#}} \\
    \multicolumn{14}{c}{\textit{Ours with Open-source and Trained LLMs}} \\
PrefRAG $_{\text{Llama3.1-8B-Instruct}}$ & 42.0 & 51.1 & 38.8 & 44.0 & 42.0 & 43.2 & 35.8 & 40.3 & 15.4 & 21.0 & 12.8 & 16.4 & 89.6 \\
PrefRAG $_{\text{GLM4-9B-chat}}$ & 45.4 & 56.3 & 42.2 & 48.0 & 55.0 & 53.7 & 42.0 & 50.2 & 23.0 & 29.4 & 20.0 & 24.1 & 87.6 \\
PrefRAG-DPO $_{\text{GLM4-9B-chat}}$ & 51.4 & 57.0 & 45.0 & 51.1 & 57.0 & 56.0 & 45.2 & 52.7 & 24.2 & 30.0 & 20.2 & 24.8 & 89.6 \\
PrefRAG $_{\text{Llama3.1-70B-Instruct}}$ & 53.6 & 63.8 & 51.8 & 56.4 & 67.4 & 66.0 & 56.8 & 63.4 & 27.0 & 34.3 & 24.2 & 28.5 & 93.2 \\
\hdashline
\multicolumn{14}{c}{\textit{Ours with Proprietary LLMs}} \\
PrefRAG $_{\text{GPT-4o-mini}}$ & 58.6 & 66.0 & 50.4 & 56.6 & 76.2 & 72.1 & 59.4 & 69.2 & 28.2 & 34.3 & 21.2 & 27.9 & 92.8 \\
$\Delta_{\text{ GPT-4o-mini}\rightarrow \text{Vanilla}\,\,\textit{w/ LR}}$ & 13.6$\uparrow$ & 12.2$\uparrow$ & 9.2$\uparrow$ & 9.9$\uparrow$ & 36.0$\uparrow$ & 27.9$\uparrow$ & 20.8$\uparrow$ & 28.2$\uparrow$ & 17.0$\uparrow$ & 15.1$\uparrow$ & 12.4$\uparrow$ & 14.8$\uparrow$ & 3.2$\uparrow$ \\
$\Delta_{\text{ GPT-4o-mini}\rightarrow \text{Vanilla${_{\textit{Mix}}}$}\,\,\textit{w/ LR $\oplus $ WR}}$ & 11.2$\uparrow$ & 8.0$\uparrow$ & 5.8$\uparrow$ & 6.6$\uparrow$ & 30.4$\uparrow$ & 23.0$\uparrow$ & 18.8$\uparrow$ & 24.1$\uparrow$ & 15.0$\uparrow$ & 13.0$\uparrow$ & 9.8$\uparrow$ & 12.6$\uparrow$ & 0.6$\uparrow$
\\
PrefRAG $_{\text{GLM4-Plus}}$ & \textbf{59.0} & \textbf{68.4} & \textbf{55.0} & \textbf{60.8} & \textbf{79.6} & \textbf{76.7} & \textbf{65.2} & \textbf{73.8} & \textbf{32.2} & \textbf{39.4} & \textbf{27.4} & \textbf{33.0} & \textbf{94.0} \\
$\Delta_{\text{ GLM4-Plus}\rightarrow \text{Vanilla}\,\,\textit{w/ LR}}$ & 12.6$\uparrow$ & 11.7$\uparrow$ & 9.2$\uparrow$ & 11.2$\uparrow$ & 34.0$\uparrow$ & 27.8$\uparrow$ & 22.2$\uparrow$ & 28.0$\uparrow$ & 16.8$\uparrow$ & 15.9$\uparrow$ & 13.6$\uparrow$ & 15.4$\uparrow$ & 4.2$\uparrow$ \\
$\Delta_{\text{ GLM4-Plus}\rightarrow \text{Vanilla${_{\textit{Mix}}}$}\,\,\textit{w/ LR $\oplus $ WR}}$ & 9.4$\uparrow$ & 7.3$\uparrow$ & 6.6$\uparrow$ & 7.8$\uparrow$ & 31.2$\uparrow$ & 25.0$\uparrow$ & 20.6$\uparrow$ & 25.6$\uparrow$ & 18.6$\uparrow$ & 15.5$\uparrow$ & 14.2$\uparrow$ & 16.1$\uparrow$ & 0.4$\uparrow$ \\
    \bottomrule
    \end{tabular}%
      \caption{{\bf Results (\%) of overall performance on all models and datasets.}}
  \label{tab-app:Overall Performance}%
\end{table*}%

\clearpage

\begin{table*}[ht!]
\setlength{\tabcolsep}{4.4pt}
    \fontsize{8}{9}\selectfont
  \centering
  
    \begin{tabular}{ccccrcccrccccc}
    \toprule
    \multirow{2}[2]{*}{\textbf{Retrieval Sources}} & \multicolumn{4}{c}{\textbf{HotpotQA}} & \multicolumn{4}{c}{\textbf{2WikiMQA}} & \multicolumn{4}{c}{\textbf{MusiQue}} & \textbf{BioASQ-Y/N} \\
\cmidrule(lr){2-5} \cmidrule(lr){6-9} \cmidrule(lr){10-13} \cmidrule(lr){14-14} & Acc. & F1 & EM & Avg. & Acc. & F1 & EM & Avg. & Acc. & F1 & EM & Avg. & Acc. \\
    \midrule
    \multicolumn{14}{c}{\textit{Llama3.1-8B-Instruct}} \\
    \hdashline
    Local Retrieval (LR) & 36.4 & 45.6 & 34.4 & 38.8 & 31.2 & 35.4 & 30.2 & 32.3 & 6.4  & 12.2 & 5.6  & 8.1  & 85.8 \\
    Web Retrieval (WR) & 36.0 & 45.1 & 34.2 & 38.4 & 28.6 & 31.4 & 24.0 & 28.0 & 5.4  & 10.8 & 4.6  & 6.9  & 87.6 \\
    LR $\oplus $ WR & \textbf{41.6} & \textbf{53.9} & \textbf{41.2} & \textbf{45.6} & \textbf{35.4} & \textbf{39.3} & \textbf{32.6} & \textbf{35.8} & \textbf{9.0} & \textbf{16.0} & \textbf{8.0} & \textbf{11.0} & \textbf{89.6} \\
    \midrule
    \multicolumn{14}{c}{\textit{GLM4-9B-chat}} \\
    \hdashline
    Local Retrieval (LR) & 34.8 & 44.4 & 34.2 & 37.8 & 34.4 & 38.8 & 33.8 & 35.7 & 8.2  & 15.0 & 7.0  & 10.1 & 87.2 \\
    Web Retrieval (WR) & 39.4 & 48.0 & 35.8 & 41.1 & 34.4 & 38.8 & 32.4 & 35.2 & 5.6  & 12.8 & 4.8 & 7.7 & 87.4 \\
    LR $\oplus $ WR & \textbf{40.8} & \textbf{51.3} & \textbf{39.0} & \textbf{43.7} & \textbf{38.8} & \textbf{43.7} & \textbf{37.4} & \textbf{40.0} & \textbf{9.0} & \textbf{16.7} & \textbf{8.4} & \textbf{11.4} & \textbf{91.0} \\
    \midrule
    \multicolumn{14}{c}{\textit{Llama3.1-70B-Instruct}}\\
    \hdashline
    Local Retrieval (LR) & 42.6 & 53.4 & 42.6 & 46.2 & 45.2 & 48.2 & 43.0 & 45.5 & 11.4 & 18.4 & 10.6 & 13.5 & 89.4 \\
    Web Retrieval (WR) & 38.8 & 50.0 & 38.4 & 42.4 & 36.6 & 38.3 & 30.4 & 35.1 & 9.4  & 15.4 & 8.8  & 11.2 & \textbf{89.6} \\
    LR $\oplus $ WR & \textbf{47.2} & \textbf{59.9} & \textbf{46.8} & \textbf{51.3} & \textbf{49.6} & \textbf{54.0} & \textbf{47.0} & \textbf{50.2} & \textbf{13.4} & \textbf{21.4} & \textbf{12.6} & \textbf{15.8} & 93.2 \\
    \midrule
    \multicolumn{14}{c}{\textit{GPT-4o-mini}} \\
    \hdashline
    Local Retrieval (LR) & 45.0 & 53.8 & 41.2 & 46.7 & 40.2 & 44.2 & 38.6 & 41.0 & 11.2 & 19.2 & 8.8  & 13.1 &  89.6 \\
    Web Retrieval (WR) & 43.4 & 53.4 & 41.0 & 45.9 & 34.4 & 39.8 & 31.0 & 35.1 & 10.2 & 17.7 & 9.2 & 12.4 & 90.2 \\
    LR $\oplus $ WR & \textbf{47.4} & \textbf{58.0} & \textbf{44.6} & \textbf{50.0} & \textbf{45.8} & \textbf{49.1} & \textbf{40.6} & \textbf{45.2} & \textbf{13.2} & \textbf{21.3} & \textbf{11.4} & \textbf{15.3} & \textbf{92.2} \\
    \midrule
    \multicolumn{14}{c}{\textit{GLM4-Plus}} \\
    \hdashline
    Local Retrieval (LR) & 46.4 & 56.7 & 45.8 & 49.6 & 45.6 & 48.9 & 43.0 & 45.8 & \textbf{15.4} & 23.5 & \textbf{13.8} & \textbf{17.6} & 89.8  \\
    Web Retrieval (WR) & 45.8 & 55.6 & 43.4 & 48.3 & 39.2 & 42.9 & 36.2 & 39.4 & 11.4 & 18.6 & 10.6 & 13.5 & 91.8  \\
    LR $\oplus $ WR & \textbf{49.6} & \textbf{61.1} & \textbf{48.4} & \textbf{53.0} & \textbf{48.4} & \textbf{51.7} & \textbf{44.6} & \textbf{48.2} & 13.6 & \textbf{23.9} & 13.2 & 16.9 & \textbf{93.6}  \\
    \bottomrule
    \end{tabular}%
    \caption{{\bf All results (\%) of Vanilla with different retrieval sources.}}
    \label{tab:All results of Vanilla with different retrieval sources} 
\end{table*}%

\begin{table*}[ht!]
\setlength{\tabcolsep}{5.2pt}
    \fontsize{8}{9}\selectfont
  \centering

    \begin{tabular}{cccccccccccccc}
    \toprule
    \multirow{2}[2]{*}{\textbf{Strategies}} & \multicolumn{4}{c}{\textbf{HotpotQA}} & \multicolumn{4}{c}{\textbf{2WikiMQA}} & \multicolumn{4}{c}{\textbf{MusiQue}} & \textbf{BioASQ-Y/N} \\
\cmidrule(lr){2-5}\cmidrule(lr){6-9}\cmidrule(lr){10-13}\cmidrule(lr){14-14} & Acc. & F1 & EM & Avg. & Acc. & F1 & EM & Avg. & Acc. & F1 & EM & Avg. & Acc. \\
    \midrule
    \multicolumn{14}{c}{\textit{Llama-3.1-8B-Instruct}} \\
    \hdashline
    PrefRAG $_{\textit{Direct}}$ & 40.6 & 48.0 & 37.0 & 41.9 & 38.8 & 40.1 & 32.0 & 37.0 & 12.2 & 17.8 & 10.2 & 13.4 & 87.0  \\
    \textbf{PrefRAG} & \textbf{42.0} & \textbf{51.1} & \textbf{38.8} & \textbf{44.0} & \textbf{42.0} & \textbf{43.2} & \textbf{35.8} & \textbf{40.3} & \textbf{15.4} & \textbf{21.0} & \textbf{12.8} & \textbf{16.4} & \textbf{89.6}  \\
    \midrule
    \multicolumn{14}{c}{\textit{GLM4-9B-chat}} \\
    \hdashline
    PrefRAG $_{\textit{Direct}}$ & 45.2 & 50.8 & 37.8 & 44.6 & 51.2 & 49.9 & 38.6 & 46.6 & 14.8 & 21.6 & 12.4 & 16.3 & \textbf{88.8} \\
    \textbf{PrefRAG} & \textbf{45.4} & \textbf{56.3} & \textbf{42.2} & \textbf{48.0} & \textbf{55.0} & \textbf{53.7} & \textbf{42.0} & \textbf{50.2} & \textbf{23.0} & \textbf{29.4} & \textbf{20.0} & \textbf{24.1} & 87.6 \\
    \midrule
    \multicolumn{14}{c}{\textit{Llama-3.1-70B-Instruct}} \\
    \hdashline
    PrefRAG $_{\textit{Direct}}$ & 50.4 & 62.4 & 49.6 & 54.1 & 61.4 & 62.5 & 54.8 & 59.6 & 23.0 & 29.9 & 21.2 & 24.7 & \textbf{93.4} \\
    \textbf{PrefRAG} & \textbf{53.6} & \textbf{63.8} & \textbf{51.8} & \textbf{56.4} & \textbf{67.4} & \textbf{66.0} & \textbf{56.8} & \textbf{63.4} & \textbf{27.0} & \textbf{34.3} & \textbf{24.2} & \textbf{28.5} & 93.2 \\
    \midrule
    \multicolumn{14}{c}{\textit{GPT-4o-mini}} \\
    \hdashline
    PrefRAG $_{\textit{Direct}}$ & 55.8 & 63.3 & 49.4 & 56.2 & \textbf{76.2} & 71.9 & 59.4 & 69.2 & 28.4 & \textbf{34.3} & 20.8 & 27.8 & 92.4 \\
    \textbf{PrefRAG} & \textbf{58.0} & \textbf{66.0} & \textbf{50.4} & \textbf{58.3} & \textbf{76.2} & \textbf{72.1} & \textbf{59.4} & \textbf{69.2} & \textbf{28.2} & \textbf{34.3} & \textbf{21.2} & \textbf{27.9} & \textbf{92.8}  \\
    \midrule
    \multicolumn{14}{c}{\textit{GLM4-Plus}} \\
    \hdashline
    PrefRAG $_{\textit{Direct}}$ & 56.4 & 66.3 & 52.4 & 58.4 & 75.6 & 72.2 & 60.6 & 69.5 & 29.8 & 36.1 & 24.6 & 30.2 & 92.6  \\
    \textbf{PrefRAG} & \textbf{59.0} & \textbf{68.4} & \textbf{55.0} & \textbf{60.8} & \textbf{79.6} & \textbf{76.7} & \textbf{65.2} & \textbf{73.8} & \textbf{32.2} & \textbf{39.4} & \textbf{27.4} & \textbf{33.0} & \textbf{94.0}  \\
    \bottomrule
    \end{tabular}%
    \caption{{\bf Results (\%) of different PrefRAG strategies.} }
    \label{tab-app:Different Strategies of PrefRAG}
\end{table*}%
\clearpage

\begin{table*}
    \fontsize{8}{9}\selectfont
  \centering
  
    \begin{tabular}{cccccc}
    \toprule
    {\multirow{2}[0]{*}{\textbf{Methods}}} & \multicolumn{5}{c}{\textbf{HotpotQA (Count)}} \\
\cmidrule(lr){2-6} & Total Local Num & Total Web Num & Total Num & Used Local Num & Used Num
\\
    \midrule
    \multicolumn{6}{c}{\textit{Llama3.1-8B-Instruct}} \\
    \hdashline
   ReAct \textit{w/} LR $\oplus$ WR & 1340 & 1340 & 2680 & 1340 & 2680\\
\textbf{PrefRAG} & 1025 & 347 & 1372 & 736 & 1083 \\
$\Delta_{\text{ Retrieval Counts}}$ & \textcolor{blue}{315$\downarrow$} & \textcolor{blue}{993$\downarrow$} & \textcolor{blue}{\textbf{1308}$\downarrow$} & \textcolor{blue}{\textbf{604}$\downarrow$} & \textcolor{blue}{\textbf{1597}$\downarrow$}\\
    \midrule
    \multicolumn{6}{c}{\textit{GLM4-9B-chat}} \\
    \hdashline
ReAct \textit{w/} LR $\oplus$ WR & 1110 & 1110 & 2220 & 1110 & 2220 \\
\textbf{PrefRAG} & 1274 & 480 & 1754 & 794 & 1274 \\
$\Delta_{\text{ Retrieval Counts}}$ & 164$\uparrow$ & \textcolor{blue}{630$\downarrow$} & \textcolor{blue}{\textbf{466}$\downarrow$} & \textcolor{blue}{\textbf{316}$\downarrow$} & \textcolor{blue}{\textbf{946}$\downarrow$} \\
\textbf{PrefRAG+DPO} & 1308 & 579 & 1887 & 729 & 1308 \\
$\Delta_{\text{ Retrieval Counts}}$ & 198$\uparrow$ & \textcolor{blue}{531$\downarrow$} & \textcolor{blue}{\textbf{333}$\downarrow$} & \textcolor{blue}{\textbf{381}$\downarrow$} & \textcolor{blue}{\textbf{912}$\downarrow$} \\
    \midrule
    \multicolumn{6}{c}{\textit{Llama3.1-70B-Instruct}} \\
    \hdashline
    ReAct \textit{w/} LR $\oplus$ WR & 930 & 930 & 1860 & 930 & 1860 \\
\textbf{PrefRAG} & 1025 & 289 & 1314 & 736 & 1025 \\
$\Delta_{\text{ Retrieval Counts}}$ & 95$\uparrow$ & \textcolor{blue}{641$\downarrow$} & \textcolor{blue}{\textbf{546}$\downarrow$} & \textcolor{blue}{\textbf{194}$\downarrow$} & \textcolor{blue}{\textbf{835}$\downarrow$} \\
    \midrule
    \multicolumn{6}{c}{\textit{GPT-4o-mini}} \\
    \midrule
    ReAct \textit{w/} LR $\oplus$ WR & 1040 & 1040 & 2080 & 1040 & 2080 \\
\textbf{PrefRAG} & 1113 & 371 & 1484 & 742 & 1113 \\
    $\Delta_{\text{ Retrieval Counts}}$ & 73$\uparrow$ & \textcolor{blue}{669$\downarrow$} & \textcolor{blue}{\textbf{596}$\downarrow$} & \textcolor{blue}{\textbf{298}$\downarrow$} & \textcolor{blue}{\textbf{967}$\downarrow$} \\
    \midrule
    \multicolumn{6}{c}{\textit{GLM4-Plus}} \\
    \hdashline
    ReAct \textit{w/} LR $\oplus$ WR & 794 & 794 & 1588 & 794 & 1588 \\
    \textbf{PrefRAG} & 1031 & 248 & 1279 & 783 & 1031 \\
    $\Delta_{\text{ Retrieval Counts}}$ & 237$\uparrow$ & \textcolor{blue}{546$\downarrow$} & \textcolor{blue}{\textbf{309}$\downarrow$} & \textcolor{blue}{\textbf{11}$\downarrow$} & \textcolor{blue}{\textbf{557}$\downarrow$} \\
    \bottomrule
    \end{tabular}%
        \caption{{\bf Total retrieval counts on HotpotQA dataset.}}
    \label{tab-app:Total counts of retrieval on HotpotQA}
\end{table*}

\begin{table*}[ht!]
    \fontsize{8}{9}\selectfont
  \centering
  
    \begin{tabular}{cccccc}
    \Xhline{1pt}
    \multirow{3}[3]{*}{\textbf{Methods}} & \multicolumn{5}{c}{\textbf{HotpotQA}} \\
    \cmidrule(lr){2-6} & \multicolumn{3}{c}{Performance (\%) ($\uparrow$)} & \multicolumn{2}{c}{Counts of Retrieval ($\downarrow$)}
    \\
\cmidrule(lr){2-4} \cmidrule(lr){5-6}       & Acc. & F1 & EM & Total Num & Used Num \\
    \hline
    \multicolumn{6}{c}{\textit{Llama3.1-8B-Instruct}} \\
    \hdashline
    ReAct \textit{w/} LR $\oplus$ WR & 41.8  & 52.0    & 39.0    & 2680  & 2680  \\
    \textbf{PrefRAG} & \cellcolor[rgb]{ .89,  .949,  .851}\textbf{42.0} & 51.1  & 38.8  & \cellcolor[rgb]{ .89,  .949,  .851}\textbf{1372 } & \cellcolor[rgb]{ .89,  .949,  .851}\textbf{1083 } \\
    \hline
    \multicolumn{6}{c}{\textit{GLM4-9B-chat}} \\
    \hdashline
    ReAct \textit{w/} LR $\oplus$ WR & 48.4  & 56.0    & 42.6  & 2220  & 2220  \\
    \textbf{PrefRAG} & \cellcolor[rgb]{ .89,  .949,  .851}45.4 & \cellcolor[rgb]{ .89,  .949,  .851}56.3 & 42.2  & \cellcolor[rgb]{ .89,  .949,  .851}\textbf{1754} & \cellcolor[rgb]{ .89,  .949,  .851}\textbf{1274} \\
    \textbf{PrefRAG+DPO} & \cellcolor[rgb]{ .89,  .949,  .851}\textbf{51.4} & \cellcolor[rgb]{ .89,  .949,  .851}\textbf{57.0} & \cellcolor[rgb]{ .89,  .949,  .851}\textbf{45.0} & \cellcolor[rgb]{ .89,  .949,  .851}1887  & \cellcolor[rgb]{ .89,  .949,  .851}1308  \\
    \hline
    \multicolumn{6}{c}{\textit{Llama3.1-70B-Instruct}} \\
    \hdashline
    ReAct \textit{w/} LR $\oplus$ WR & 51.6  & 63.7  & 50.6  & 1860  & 1860  \\
    \textbf{PrefRAG} & \cellcolor[rgb]{ .89,  .949,  .851}\textbf{53.6} & \cellcolor[rgb]{ .89,  .949,  .851}\textbf{63.8} & \cellcolor[rgb]{ .89,  .949,  .851}\textbf{51.8} & \cellcolor[rgb]{ .89,  .949,  .851}\textbf{1314} & \cellcolor[rgb]{ .89,  .949,  .851}\textbf{1025} \\
    \hline
    \multicolumn{6}{c}{\textit{GPT-4o-mini}} \\
    \hdashline
    ReAct \textit{w/} LR $\oplus$ WR & 57.0 & 65.9  & 51.4  & 2080  & 2080  \\
    \textbf{PrefRAG} & \cellcolor[rgb]{ .89,  .949,  .851}\textbf{58.6} & \cellcolor[rgb]{ .89,  .949,  .851}\textbf{66.0} & 50.4  & \cellcolor[rgb]{ .89,  .949,  .851}\textbf{1484} & \cellcolor[rgb]{ .89,  .949,  .851}\textbf{1113} \\
    \hline
    \multicolumn{6}{c}{\textit{GLM4-Plus}} \\
    \hdashline
    ReAct \textit{w/} LR $\oplus$ WR & \textcolor[rgb]{ .122,  .137,  .161}{56.6} & \textcolor[rgb]{ .122,  .137,  .161}{67.0} & \textcolor[rgb]{ .122,  .137,  .161}{53.6} & 1588  & 1588 \\
    \textbf{PrefRAG} & \cellcolor[rgb]{ .89,  .949,  .851}\textbf{59.0} & \cellcolor[rgb]{ .89,  .949,  .851}\textbf{68.4} & \cellcolor[rgb]{ .89,  .949,  .851}\textbf{55.0} & \cellcolor[rgb]{ .89,  .949,  .851}\textbf{1279} & \cellcolor[rgb]{ .89,  .949,  .851}\textbf{1031} \\
    \Xhline{1pt}
    \end{tabular}%
    \caption{{\bf Efficiency and accuracy trade-off on HotpotQA dataset.} }
    \label{tab-app:Efficiency and accuracy trade-off on HotpotQA}
\end{table*}%

\begin{table*}
    \fontsize{8}{9}\selectfont
  \centering
  
    \begin{tabular}{cccccc}
    \toprule
    {\multirow{2}[2]{*}{\textbf{Methods}}} & \multicolumn{5}{c}{\textbf{2WikiMQA (Count)}}\\
\cmidrule(lr){2-6} & Total Local Num & Total Web Num & Total Num & Used Local Num & Used Num \\
    \midrule
    \multicolumn{6}{c}{\textit{Llama3.1-8B-Instruct}} \\
    \hdashline
   ReAct \textit{w/} LR $\oplus$ WR & 1207 & 1207 & 2414 & 1207 & 2414 \\
\textbf{PrefRAG} & 1134 & 513 & 1647 & 623 & 1136 \\
$\Delta_{\text{ Retrieval Counts}}$ & \textcolor{blue}{73$\downarrow$} & \textcolor{blue}{694$\downarrow$} & \textcolor{blue}{\textbf{767}$\downarrow$} & \textcolor{blue}{\textbf{584}$\downarrow$} & \textcolor{blue}{\textbf{1278}$\downarrow$} \\
    \midrule
    \multicolumn{6}{c}{\textit{GLM4-9B-chat}} \\
    \hdashline
ReAct \textit{w/} LR $\oplus$ WR & 1259 & 1259 & 2518 & 1259 & 2518 \\
\textbf{PrefRAG} & 1330 & 387 & 1717 & 943 & 1330 \\
$\Delta_{\text{ Retrieval Counts}}$ & 71$\uparrow$ & \textcolor{blue}{872$\downarrow$} & \textcolor{blue}{\textbf{801}$\downarrow$} & \textcolor{blue}{\textbf{316}$\downarrow$} & \textcolor{blue}{\textbf{1188}$\downarrow$} \\
\textbf{PrefRAG+DPO} & 1354 & 431 & 1785 & 923 & 1354 \\
$\Delta_{\text{ Retrieval Counts}}$ & 95$\uparrow$ & \textcolor{blue}{828$\downarrow$} & \textcolor{blue}{\textbf{733}$\downarrow$} & \textcolor{blue}{\textbf{336}$\downarrow$} & \textcolor{blue}{\textbf{1164}$\downarrow$} \\
    \midrule
    \multicolumn{6}{c}{\textit{Llama3.1-70B-Instruct}} \\
    \hdashline
    ReAct \textit{w/} LR $\oplus$ WR & 1189 & 1189 & 2378 & 1189 & 2378 \\
\textbf{PrefRAG} & 1132 & 271 & 1403 & 861 & 1132 \\
$\Delta_{\text{ Retrieval Counts}}$ & \textcolor{blue}{57$\downarrow$} & \textcolor{blue}{918$\downarrow$} & \textcolor{blue}{\textbf{975}$\downarrow$} & \textcolor{blue}{\textbf{328}$\downarrow$} & \textcolor{blue}{\textbf{1246}$\downarrow$} \\
    \midrule
    \multicolumn{6}{c}{\textit{GPT-4o-mini}} \\
    \midrule
    ReAct \textit{w/} LR $\oplus$ WR & 1302 & 1302 & 2604 & 1302 & 2604 \\
\textbf{PrefRAG} & 1357 & 485 & 1842 & 872 & 1357 \\
    $\Delta_{\text{ Retrieval Counts}}$ & 55$\uparrow$ & \textcolor{blue}{817$\downarrow$} & \textcolor{blue}{\textbf{762}$\downarrow$} & \textcolor{blue}{\textbf{430}$\downarrow$} & \textcolor{blue}{\textbf{1247}$\downarrow$} \\
    \midrule
    \multicolumn{6}{c}{\textit{GLM4-Plus}} \\
    \hdashline
    ReAct \textit{w/} LR $\oplus$ WR & 913 & 913 & 1826 & 913 & 1826 \\
    \textbf{PrefRAG} & 1200 & 248 & 1448 & 952 & 1200 \\
    $\Delta_{\text{ Retrieval Counts}}$ & 287$\uparrow$ & \textcolor{blue}{665$\downarrow$} & \textcolor{blue}{\textbf{378}$\downarrow$} & \textbf{39}$\uparrow$ & \textcolor{blue}{\textbf{626}$\downarrow$} \\
    \bottomrule
    \end{tabular}%
        \caption{{\bf Total retrieval counts on 2WikiMQA dataset.} }
    \label{tab-app:Total counts of retrieval on 2WikiMQA}
\end{table*}

\begin{table*}[ht!]
    \fontsize{8}{9}\selectfont
  \centering
  
    \begin{tabular}{cccccc}
    \Xhline{1pt}
    \multirow{3}[3]{*}{\textbf{Methods}} & \multicolumn{5}{c}{\textbf{2WikiMQA}} \\
    \cmidrule(lr){2-6} & \multicolumn{3}{c}{Performance (\%) ($\uparrow$)} & \multicolumn{2}{c}{Counts of Retrieval ($\downarrow$)}
    \\
\cmidrule(lr){2-4} \cmidrule(lr){5-6}       & Acc. & F1 & EM & Total Num & Used Num \\
    \hline
    \multicolumn{6}{c}{\textit{Llama3.1-8B-Instruct}} \\
    \hdashline
    ReAct \textit{w/} LR $\oplus$ WR & 38.0  & 39.4  & 30.6  & 2414  & 2414  \\
    \textbf{PrefRAG} & \cellcolor[rgb]{ .89,  .949,  .851}\textbf{42.0} & \cellcolor[rgb]{ .89,  .949,  .851}\textbf{43.2} & \cellcolor[rgb]{ .89,  .949,  .851}\textbf{35.8} & \cellcolor[rgb]{ .89,  .949,  .851}\textbf{1647} & \cellcolor[rgb]{ .89,  .949,  .851}\textbf{1136} \\
    \hline
    \multicolumn{6}{c}{\textit{GLM4-9B-chat}} \\
    \hdashline
    ReAct \textit{w/} LR $\oplus$ WR & 56.8 & 54.6 & 41.2 & 2518 & 2518  \\
    \textbf{PrefRAG} & 55.0 & 53.7 & \cellcolor[rgb]{ .89,  .949,  .851}42.0 & \cellcolor[rgb]{ .89,  .949,  .851}\textbf{1717} & \cellcolor[rgb]{ .89,  .949,  .851}\textbf{1330} \\
    \textbf{PrefRAG+DPO} & \cellcolor[rgb]{ .89,  .949,  .851}\textbf{57.0} & \cellcolor[rgb]{ .89,  .949,  .851}\textbf{56.0} & \cellcolor[rgb]{ .89,  .949,  .851}\textbf{45.2} & \cellcolor[rgb]{ .89,  .949,  .851}1785 & \cellcolor[rgb]{ .89,  .949,  .851}1354 \\
    \hline
    \multicolumn{6}{c}{\textit{Llama3.1-70B-Instruct}} \\
    \hdashline
    ReAct \textit{w/} LR $\oplus$ WR & 68.2  & 68.7  & 61.4  & 2378  & 2378  \\
    \textbf{PrefRAG} & 67.4  & 66.0  & 56.8  & \cellcolor[rgb]{ .89,  .949,  .851}\textbf{1403} & \cellcolor[rgb]{ .89,  .949,  .851}\textbf{1132} \\
    \hline
    \multicolumn{6}{c}{\textit{GPT-4o-mini}} \\
    \hdashline
    ReAct \textit{w/} LR $\oplus$ WR & 78.4  & 74.1  & 61.8  & 2604  & 2604  \\
    \textbf{PrefRAG} & 76.2 & 72.1 & 59.4 & \cellcolor[rgb]{ .89,  .949,  .851}\textbf{1842} & \cellcolor[rgb]{ .89,  .949,  .851}\textbf{1357} \\
    \hline
    \multicolumn{6}{c}{\textit{GLM4-Plus}} \\
    \hdashline
    ReAct \textit{w/} LR $\oplus$ WR & \textcolor[rgb]{ .122,  .137,  .161}{73.8} & \textcolor[rgb]{ .122,  .137,  .161}{70.5} & \textcolor[rgb]{ .122,  .137,  .161}{59.0} & 1826 & 1826 \\
    \textbf{PrefRAG} & \cellcolor[rgb]{ .89,  .949,  .851}\textbf{79.6} & \cellcolor[rgb]{ .89,  .949,  .851}\textbf{76.7} & \cellcolor[rgb]{ .89,  .949,  .851}\textbf{65.2} & \cellcolor[rgb]{ .89,  .949,  .851}\textbf{1448} & \cellcolor[rgb]{ .89,  .949,  .851}\textbf{1200} \\
    \Xhline{1pt}
    \end{tabular}%
    \caption{{\bf Efficiency and accuracy trade-off on 2WikiMQA dataset.} }
    \label{tab-app:Efficiency and accuracy trade-off on 2WikiMQA}
\end{table*}%

\begin{table*}
    \fontsize{8}{9}\selectfont
  \centering
  
    \begin{tabular}{cccccc}
    \toprule
    {\multirow{2}[0]{*}{\textbf{Methods}}} & \multicolumn{5}{c}{\textbf{MusiQue (Count)}} \\
\cmidrule(lr){2-6} & Total Local Num & Total Web Num & Total Num & Used Local Num & Used Num \\
    \midrule
    \multicolumn{6}{c}{\textit{Llama3.1-8B-Instruct}} \\
    \hdashline
   ReAct \textit{w/} LR $\oplus$ WR & 1444 & 1444 & 2888 & 1444 & 2888 \\
\textbf{PrefRAG} & 1369 & 695 & 2064 & 675 & 1370 \\
$\Delta_{\text{ Retrieval Counts}}$ & \textcolor{blue}{75$\downarrow$} & \textcolor{blue}{749$\downarrow$} & \textcolor{blue}{\textbf{824}$\downarrow$} & \textcolor{blue}{\textbf{769}$\downarrow$} & \textcolor{blue}{\textbf{1518}$\downarrow$} \\
    \midrule
    \multicolumn{6}{c}{\textit{GLM4-9B-chat}} \\
    \hdashline
ReAct \textit{w/} LR $\oplus$ WR & 1478 & 1478 & 2956 & 1478 & 2956 \\
\textbf{PrefRAG} & 1625 & 835 & 2460 & 790 & 1625 \\
$\Delta_{\text{ Retrieval Counts}}$ & 147$\uparrow$ & \textcolor{blue}{643$\downarrow$} & \textcolor{blue}{\textbf{496}$\downarrow$} & \textcolor{blue}{\textbf{688}$\downarrow$} & \textcolor{blue}{\textbf{1331}$\downarrow$} \\
\textbf{PrefRAG+DPO} & 1643 & 996 & 2639 & 647 & 1643 \\
$\Delta_{\text{ Retrieval Counts}}$ & 165$\uparrow$ & \textcolor{blue}{482$\downarrow$} & \textcolor{blue}{\textbf{317}$\downarrow$} & \textcolor{blue}{\textbf{831}$\downarrow$} & \textcolor{blue}{\textbf{1313}$\downarrow$} \\
    \midrule
    \multicolumn{6}{c}{\textit{Llama3.1-70B-Instruct}} \\
    \hdashline
    ReAct \textit{w/} LR $\oplus$ WR & 1241 & 1241 & 2482 & 1241 & 2482 \\
\textbf{PrefRAG} & 1170 & 452 & 1622 & 718 & 1170 \\
$\Delta_{\text{ Retrieval Counts}}$ & \textcolor{blue}{71$\downarrow$} & \textcolor{blue}{789$\downarrow$} & \textcolor{blue}{\textbf{860}$\downarrow$} & \textcolor{blue}{\textbf{523}$\downarrow$} & \textcolor{blue}{\textbf{1312}$\downarrow$} \\
    \midrule
    \multicolumn{6}{c}{\textit{GPT-4o-mini}} \\
    \midrule
    ReAct \textit{w/} LR $\oplus$ WR & 1515 & 1515 & 3030 & 1515 & 3030 \\
\textbf{PrefRAG} & 1562 & 885 & 2447 & 677 & 1562 \\
    $\Delta_{\text{ Retrieval Counts}}$ & 47$\uparrow$ & \textcolor{blue}{630$\downarrow$} & \textcolor{blue}{\textbf{583}$\downarrow$} & \textcolor{blue}{\textbf{838}$\downarrow$} & \textcolor{blue}{\textbf{1468}$\downarrow$} \\
    \midrule
    \multicolumn{6}{c}{\textit{GLM4-Plus}} \\
    \hdashline
    ReAct \textit{w/} LR $\oplus$ WR & 918 & 918 & 1836 & 918 & 1836 \\
    \textbf{PrefRAG} & 1373 & 603 & 1976 & 770 & 1373 \\
    $\Delta_{\text{ Retrieval Counts}}$ & 455$\uparrow$ & \textcolor{blue}{315$\downarrow$} & \textbf{140}$\uparrow$ & \textcolor{blue}{\textbf{148}$\downarrow$} & \textcolor{blue}{\textbf{463}$\downarrow$} \\
    \bottomrule
    \end{tabular}%
        \caption{{\bf Total retrieval counts on MusiQue dataset.} }
    \label{tab-app:Total counts of retrieval on MusiQue}
\end{table*}

\begin{table*}[ht!]
    \fontsize{8}{9}\selectfont
  \centering
  
    \begin{tabular}{cccccc}
    \Xhline{1pt}
    \multirow{3}[3]{*}{\textbf{Methods}} & \multicolumn{5}{c}{\textbf{MusiQue}} \\
    \cmidrule(lr){2-6} & \multicolumn{3}{c}{Performance (\%) ($\uparrow$)} & \multicolumn{2}{c}{Counts of Retrieval ($\downarrow$)}
    \\
\cmidrule(lr){2-4} \cmidrule(lr){5-6}       & Acc. & F1 & EM & Total Num & Used Num \\
    \hline
    \multicolumn{6}{c}{\textit{Llama3.1-8B-Instruct}} \\
    \hdashline
    ReAct \textit{w/} LR $\oplus$ WR & 12.8 & 19.3 & 10.4 & \multicolumn{1}{c}{2888} & \multicolumn{1}{c}{2888} \\
    \textbf{PrefRAG} & \cellcolor[rgb]{ .89,  .949,  .851}\textbf{15.4} & \cellcolor[rgb]{ .89,  .949,  .851}\textbf{21.0 } & \cellcolor[rgb]{ .89,  .949,  .851}\textbf{12.8} & \multicolumn{1}{c}{\cellcolor[rgb]{ .89,  .949,  .851}\textbf{2064}} & \multicolumn{1}{c}{\cellcolor[rgb]{ .89,  .949,  .851}\textbf{1370}} \\
    \hline
    \multicolumn{6}{c}{\textit{GLM4-9B-chat}} \\
    \hdashline
    ReAct \textit{w/} LR $\oplus$ WR & 22.0  & 28.7 & 18.8 & \multicolumn{1}{c}{2956} & \multicolumn{1}{c}{2956} \\
    \textbf{PrefRAG} & \cellcolor[rgb]{ .89,  .949,  .851}23.0 & \cellcolor[rgb]{ .89,  .949,  .851}29.4 & \cellcolor[rgb]{ .89,  .949,  .851}20.0 & \multicolumn{1}{c}{\cellcolor[rgb]{ .89,  .949,  .851}\textbf{2460}} & \multicolumn{1}{c}{\cellcolor[rgb]{ .89,  .949,  .851}\textbf{1625}} \\
    \textbf{PrefRAG+DPO} & \cellcolor[rgb]{ .89,  .949,  .851}\textbf{24.2} & \cellcolor[rgb]{ .89,  .949,  .851}\textbf{30.0} & \cellcolor[rgb]{ .89,  .949,  .851}\textbf{20.2} & \multicolumn{1}{c}{\cellcolor[rgb]{ .89,  .949,  .851}2639} & \multicolumn{1}{c}{\cellcolor[rgb]{ .89,  .949,  .851}1643} \\
    \midrule
    \multicolumn{6}{c}{\textit{Llama3.1-70B-Instruct}} \\
    \hdashline
    ReAct \textit{w/} LR $\oplus$ WR & 25.0  & 34.0  & 23.8  & \multicolumn{1}{c}{2482} & \multicolumn{1}{c}{2482} \\
    \textbf{PrefRAG} & \cellcolor[rgb]{ .89,  .949,  .851}\textbf{27.0} & \cellcolor[rgb]{ .89,  .949,  .851}\textbf{34.3} & \cellcolor[rgb]{ .89,  .949,  .851}\textbf{24.2} & \multicolumn{1}{c}{\cellcolor[rgb]{ .89,  .949,  .851}\textbf{1622}} & \multicolumn{1}{c}{\cellcolor[rgb]{ .89,  .949,  .851}\textbf{1170}} \\
    \hline
    \multicolumn{6}{c}{\textit{GPT-4o-mini}} \\
    \hdashline
    ReAct \textit{w/} LR $\oplus$ WR & 28.4  & 34.8  & 21.4  & \multicolumn{1}{c}{3030} & \multicolumn{1}{c}{3030} \\
    \textbf{PrefRAG} & 28.2 & 34.3 & 21.2 & \multicolumn{1}{c}{\cellcolor[rgb]{ .89,  .949,  .851}\textbf{2447}} & \multicolumn{1}{c}{\cellcolor[rgb]{ .89,  .949,  .851}\textbf{1562}} \\
    \hline
    \multicolumn{6}{c}{\textit{GLM4-Plus}} \\
    \hdashline
    ReAct \textit{w/} LR $\oplus$ WR & \textcolor[rgb]{ .122,  .137,  .161}{25.8} & \textcolor[rgb]{ .122,  .137,  .161}{33.3} & \textcolor[rgb]{ .122,  .137,  .161}{21.2} & 1836  & 1836 \\
    \textbf{PrefRAG} & \cellcolor[rgb]{ .89,  .949,  .851}\textbf{32.2} & \cellcolor[rgb]{ .89,  .949,  .851}\textbf{39.4} & \cellcolor[rgb]{ .89,  .949,  .851}\textbf{27.4} & 1976 & \cellcolor[rgb]{ .89,  .949,  .851}\textbf{1373} \\
    \Xhline{1pt}
    \end{tabular}%
    \caption{{\bf Efficiency and accuracy trade-off on MusiQue dataset.} }
    \label{tab-app:Efficiency and accuracy trade-off on MusiQue}
\end{table*}%
\clearpage

\begin{table*}
    \fontsize{8}{9}\selectfont
  \centering
  
    \begin{tabular}{cccccc}
    \toprule
    {\multirow{2}[0]{*}{\textbf{Methods}}} & \multicolumn{5}{c}{\textbf{BioASQ-Y/N (Count)}} \\
\cmidrule(lr){2-6} & Total Local Num & Total Web Num & Total Num & Used Local Num & Used Num \\
    \midrule
    \multicolumn{6}{c}{\textit{Llama3.1-8B-Instruct}} \\
    \hdashline
   ReAct \textit{w/} LR $\oplus$ WR & 1205 & 1205 & 2410 & 1205 & 2410 \\
    \textbf{PrefRAG} & 1178 & 210 & 1388 & 968 & 1178 \\
    $\Delta_{\text{ Retrieval Counts}}$ & \textcolor{blue} 
    {\textbf{27}$\downarrow$} & \textcolor{blue}{995$\downarrow$} & \textcolor{blue}{\textbf{1022}$\downarrow$} & \textcolor{blue}{\textbf{237}$\downarrow$} & \textcolor{blue}{\textbf{1232}$\downarrow$} \\
    \midrule
    \multicolumn{6}{c}{\textit{GLM4-9B-chat}} \\
    \hdashline
    ReAct \textit{w/} LR $\oplus$ WR & 829 & 829 & 1658 & 829 & 1658 \\
    \textbf{PrefRAG} & 1116 & 213 & 1329 & 934 &1147 \\
    $\Delta_{\text{ Retrieval Counts}}$ & 287$\uparrow$ & \textcolor{blue}{616$\downarrow$} & \textcolor{blue}{\textbf{329}$\downarrow$} & 105$\uparrow$ & \textcolor{blue}{\textbf{511}$\downarrow$} \\
    \textbf{PrefRAG+DPO} & 1383 & 726 & 2109 & 657 & 1383 \\
    $\Delta_{\text{ Retrieval Counts}}$ & 554$\uparrow$ & \textcolor{blue}{103$\downarrow$} & 451$\uparrow$ & \textcolor{blue}{\textbf{172}$\downarrow$} & \textcolor{blue}{\textbf{275}$\downarrow$} \\
    \midrule
    \multicolumn{6}{c}{\textit{Llama3.1-70B-Instruct}} \\
    \hdashline
    ReAct \textit{w/} LR $\oplus$ WR & 1097 & 1097 & 2194 & 1097 & 2194 \\
    \textbf{PrefRAG} & 1052 & 472 & 1524 & 666 & 1138 \\
    $\Delta_{\text{ Retrieval Counts}}$ & \textcolor{blue}{45$\downarrow$} & \textcolor{blue}{625$\downarrow$} & \textcolor{blue}{\textbf{670}$\downarrow$} & \textcolor{blue}{\textbf{431}$\downarrow$} & \textcolor{blue}{\textbf{1056}$\downarrow$} \\
    \midrule
    \multicolumn{6}{c}{\textit{GPT-4o-mini}} \\
    \midrule
    ReAct \textit{w/} LR $\oplus$ WR & 714 & 714 & 1428 & 714 & 1428 \\
    \textbf{PrefRAG} & 799 & 202 & 1001 & 599 & 801 \\
    $\Delta_{\text{ Retrieval Counts}}$ & 85$\uparrow$ & \textcolor{blue}{512$\downarrow$} & \textcolor{blue}{\textbf{427}$\downarrow$} & \textcolor{blue}{\textbf{115}$\downarrow$} & \textcolor{blue}{\textbf{627}$\downarrow$} \\
    \midrule
    \multicolumn{6}{c}{\textit{GLM4-Plus}} \\
    \hdashline
    ReAct \textit{w/} LR $\oplus$ WR & 665 & 665 & 1330 & 665 & 1330 \\
    \textbf{PrefRAG} & 681 & 118 & 799 & 583 & 701 \\
    $\Delta_{\text{ Retrieval Counts}}$ & 16$\uparrow$ & \textcolor{blue}{547$\downarrow$} & \textcolor{blue}{\textbf{531}$\downarrow$} & \textcolor{blue}{\textbf{82}$\downarrow$} & \textcolor{blue}{\textbf{629}$\downarrow$} \\
    \bottomrule
    \end{tabular}%
        \caption{{\bf Total retrieval counts on BioASQ-Y/N dataset.}}
    \label{tab-app:Total counts of retrieval on BioASQ-Y/N}
\end{table*}

\begin{table*}[ht!]
    \fontsize{8}{9}\selectfont
  \centering
  
    \begin{tabular}{cccc}
    \Xhline{1pt}
    \multirow{3}[3]{*}{\textbf{Methods}} & \multicolumn{3}{c}{\textbf{BioASQ-Y/N}} \\
    \cmidrule(lr){2-4} & \multicolumn{1}{c}{Performance (\%) ($\uparrow$)} & \multicolumn{2}{c}{Counts of Retrieval ($\downarrow$)}
    \\
\cmidrule(lr){2-2} \cmidrule(lr){3-4}       & Acc. & Total Num & Used Num \\
    \hline
    \multicolumn{4}{c}{\textit{Llama3.1-8B-Instruct}} \\
    \hdashline
    ReAct \textit{w/} LR $\oplus$ WR & 87.8 & \multicolumn{1}{c}{2410} & \multicolumn{1}{c}{2410} \\
    \textbf{PrefRAG} & \cellcolor[rgb]{ .89,  .949,  .851}\textbf{89.6} & \multicolumn{1}{c}{\cellcolor[rgb]{ .89,  .949,  .851}\textbf{1388}} & \multicolumn{1}{c}{\cellcolor[rgb]{ .89,  .949,  .851}\textbf{1178}} \\
    \hline
    \multicolumn{4}{c}{\textit{GLM4-9B-chat}} \\
    \hdashline
    ReAct \textit{w/} LR $\oplus$ WR & 87.4 & \multicolumn{1}{c}{1658} & \multicolumn{1}{c}{1658} \\
    \textbf{PrefRAG} & \cellcolor[rgb]{ .89,  .949,  .851}87.6 & \multicolumn{1}{c}{\cellcolor[rgb]{ .89,  .949,  .851}\textbf{1329}} & \multicolumn{1}{c}{\cellcolor[rgb]{ .89,  .949,  .851}\textbf{1147}} \\
    \textbf{PrefRAG+DPO} & \cellcolor[rgb]{ .89,  .949,  .851}\textbf{89.6} & \multicolumn{1}{c}{2109} & \multicolumn{1}{c}{\cellcolor[rgb]{ .89,  .949,  .851}1383} \\
    \midrule
    \multicolumn{4}{c}{\textit{Llama3.1-70B-Instruct}} \\
    \hdashline
    ReAct \textit{w/} LR $\oplus$ WR & 93.6 & \multicolumn{1}{c}{2194} & \multicolumn{1}{c}{2194} \\
    \textbf{PrefRAG} & 93.2 & \multicolumn{1}{c}{\cellcolor[rgb]{ .89,  .949,  .851}\textbf{1524}} & \multicolumn{1}{c}{\cellcolor[rgb]{ .89,  .949,  .851}\textbf{1138}} \\
    \hline
    \multicolumn{4}{c}{\textit{GPT-4o-mini}} \\
    \hdashline
    ReAct \textit{w/} LR $\oplus$ WR & 91.4 & \multicolumn{1}{c}{1428} & \multicolumn{1}{c}{1428} \\
    \textbf{PrefRAG} & {\cellcolor[rgb]{ .89,  .949,  .851}\textbf{92.8}} & \multicolumn{1}{c}{\cellcolor[rgb]{ .89,  .949,  .851}\textbf{1001}} & \multicolumn{1}{c}{\cellcolor[rgb]{ .89,  .949,  .851}\textbf{801}} \\
    \hline
    \multicolumn{4}{c}{\textit{GLM4-Plus}} \\
    \hdashline
    ReAct \textit{w/} LR $\oplus$ WR & \textcolor[rgb]{ .122,  .137,  .161}{93.2} & 1330  & 1330 \\
    \textbf{PrefRAG} & \cellcolor[rgb]{ .89,  .949,  .851}\textbf{94.0} & \cellcolor[rgb]{ .89,  .949,  .851}\textbf{799} & \cellcolor[rgb]{ .89,  .949,  .851}\textbf{701} \\
    \Xhline{1pt}
    \end{tabular}%
    \caption{{\bf Efficiency and accuracy trade-off on BioASQ-Y/N dataset.} }
    \label{tab-app:Efficiency and accuracy trade-off on BioASQ-Y/N}
\end{table*}%
\clearpage

\begin{table*}[ht!]
\renewcommand{\arraystretch}{1.3}
\setlength{\tabcolsep}{3.5pt}
    \fontsize{7}{8}\selectfont
  \centering

    \begin{tabular}{ccccccccccccccc}
    \Xhline{1pt}
    \multirow{2}[2]{*}{\textbf{LLMs}} & \multirow{2}[2]{*}{\textbf{Methods}} & \multicolumn{4}{c}{\textbf{HotpotQA}} & \multicolumn{4}{c}{\textbf{2WikiMQA}} & \multicolumn{4}{c}{\textbf{MusiQue}} & \textbf{BioASQ-Y/N} \\
    \cmidrule(lr){3-6} \cmidrule(lr){7-10} \cmidrule(lr){11-14} \cmidrule(lr){15-15} &  & Acc.  & F1    & EM    & Avg.  & Acc.  & F1    & EM    & Avg.  & Acc.  & F1    & EM    & Avg.  & Acc. \\
    \hline
    \multirow{3}[0]{*}{Llama3.1-8B-Instruct} & \textbf{PrefRAG} &  \textbf{42.0} &  \textbf{51.1} & 38.8 & \multicolumn{1}{c}{ \textbf{44.0}} &  \textbf{42.0} & \textbf{43.2} &  \textbf{35.8} &  \textbf{40.3} &  \textbf{15.4} &  \textbf{21.0} &  \textbf{12.8} &  \textbf{16.4} & \textbf{89.6}  \\
    & \textit{w/o} Pref-AR & 41.0  & \underline{50.9}  &  \textbf{39.8} & \underline{43.9}  & 36.0  & 37.8  & 30.2  & 34.7  & \underline{13.6}  & 19.0  & 11.0  & 14.5  & \underline{81.4}  \\
    & \textit{w/o} Self-Reflection & \underline{41.6}  & \underline{50.9}  & \underline{39.6} &  \textbf{44.0} & \underline{41.6}  & \underline{42.1}  & \underline{34.4}  & \underline{39.4}  & 13.2  & \underline{19.9}  & \underline{12.2}  & \underline{15.1}  &  \textbf{89.6} \\
    \hdashline
    \multirow{3}[0]{*}{GLM4-9B-chat} & \textbf{PrefRAG} & 45.4  & \underline{56.3}  & \underline{42.2}  & \underline{48.0}  &  \textbf{55.0} & \underline{53.7}  & \underline{42.0}  & \underline{50.2}  &  \textbf{23.0} &  \textbf{29.4} &  \textbf{20.0} &  \textbf{24.1} & \underline{87.6}  \\
    & \textit{w/o} Pref-AR & \underline{46.8}  & 54.8  & \underline{42.2}  & 47.9  & 51.0  & 51.2  & 39.0  & 47.1  & 16.0  & 22.5  & 12.8  & 17.1  &  87.0 \\
    & \textit{w/o} Self-Reflection &  \textbf{47.0} &  \textbf{57.4} &  \textbf{45.0} &  \textbf{49.8} & \underline{53.8}  &  \textbf{54.3} &  \textbf{43.4} &  \textbf{50.5} & \underline{21.4}  & \underline{27.4}  & \underline{18.2}  & \underline{22.3}  & \textbf{89.0}  \\
    \hdashline
    \multirow{3}[0]{*}{GLM4-9B-chat-DPO} & \textbf{PrefRAG} &  \textbf{51.4} &  \textbf{57.0} &  \textbf{45.0} &  \textbf{51.1} &  \textbf{57.0} &  \textbf{56.0} &  \textbf{45.2} &  \textbf{52.7} &  \textbf{24.2} &  \textbf{30.0} &  \textbf{20.2} &  \textbf{24.8} &  \underline{89.6} \\
    & \textit{w/o} Pref-AR & 47.4  & 53.4  & 41.0  & 47.3  & 53.6  & 53.4  & 40.0  & 49.0  & 18.0  & 23.1  & 14.4  & 18.5  & 88.8  \\
    & \textit{w/o} Self-Reflection & \underline{49.4}  & \underline{56.0}  & \underline{42.6} & \underline{49.3}  & \underline{56.8}  & \underline{54.4}  & \underline{41.8}  & \underline{51.0}  & \underline{22.4}  & \underline{28.0}  & \underline{18.4}  & \underline{22.9}  & \textbf{89.8}  \\
    \hdashline
    \multirow{3}[0]{*}{Llama3.1-70B-Instruct} & \textbf{PrefRAG} &  \textbf{53.6} &  \textbf{63.8} &  \textbf{51.8} &  \textbf{56.4} &  \textbf{67.4} &  \underline{66.0} & \underline{56.8}  &  \textbf{63.4} &  \textbf{27.0} &  \textbf{34.3} & \underline{24.2}  &  \textbf{28.5} & \textbf{93.2}  \\
    & \textit{w/o} Pref-AR & \underline{52.6}  & \underline{63.5}  & \underline{51.4}  & \underline{55.8}  & 64.8  & 63.8  & 54.8  & 61.1  & 25.4  & 33.6  & 22.4  & \underline{27.1}  & \underline{92.4}  \\
    & \textit{w/o} Self-Reflection & 51.4  & 63.0  & 49.4  & 54.6  & \underline{66.2}  & \textbf{66.1}  &  \textbf{57.0} & \underline{63.1}  & \underline{26.8}  & \underline{34.2}  &  \textbf{24.4} & \textbf{28.5} & 92.2 \\
    \hdashline
    \multirow{3}[0]{*}{GPT-4o-mini} & \textbf{PrefRAG} & \textbf{58.6} & \underline{66.0}  & \underline{50.4}  & \underline{58.3}  & \underline{76.2}  & \textbf{72.1} & \underline{59.4}  & \underline{69.2}  & \underline{28.2}  & \textbf{34.3} & \textbf{21.2} & \textbf{27.9} & \textbf{92.8} \\
    & \textit{w/o} Pref-AR & 51.4  & 58.4  & 43.8  & 51.2  & 69.8  & 66.8  & 52.6  & 63.1  & 19.6  & 26.7  & 14.4  & 20.2  & 89.4  \\
    & \textit{w/o} Self-Reflection & \underline{57.8}  & \textbf{66.2} & \textbf{51.6} & \textbf{58.5} & \textbf{76.6} & \underline{71.9}  & \textbf{59.8} & \textbf{69.4} & \textbf{28.6} & \underline{33.7}  & \underline{21.0}  & \underline{27.8}  & \underline{92.4}  \\
    \hdashline
    \multicolumn{1}{c}{\multirow{3}[0]{*}{GLM4-Plus}} & \multicolumn{1}{c}{\textbf{PrefRAG}} & \multicolumn{1}{c}{\textbf{59.0}} & \multicolumn{1}{c}{\textbf{68.4}} & \multicolumn{1}{c}{\textbf{55.0}} & \multicolumn{1}{c}{\textbf{60.8}} & \multicolumn{1}{c}{\textbf{79.6}} & \multicolumn{1}{c}{\textbf{76.7}} & \multicolumn{1}{c}{\textbf{65.2}} & \multicolumn{1}{c}{\textbf{73.8}} & \multicolumn{1}{c}{\textbf{32.2}} & \multicolumn{1}{c}{\textbf{39.4}} & \multicolumn{1}{c}{\textbf{27.4}} & \textbf{33.0}  & \textbf{94.0} \\
    & \textit{w/o} Pref-AR & 51.6  & 61.1  & 47.8 & 53.5  & 74.2  & 72.6  & 59.6  & 68.8  & 26.2  & 33.3  & 22.0  & 27.2  & 93.4  \\
    & \textit{w/o} Self-Reflection & \underline{57.6}  & \underline{67.3}  & \underline{53.8}  & \underline{59.6}  & \underline{78.6}  & \underline{74.8}  & \underline{62.8}  & \underline{72.1}  & \underline{32.0} & \underline{38.5}  & \underline{27.0} & \underline{32.5} & \underline{93.6}  \\
    \Xhline{1pt}
    \end{tabular}%
    \caption{{\bf All results (\%) of ablation study.} }
    \label{tab-app:All ablation results}
\end{table*}%

\begin{table*}[ht!]
\setlength{\tabcolsep}{4pt}
    \fontsize{8}{9}\selectfont
  \centering

    \begin{tabular}{ccccccccccccccc}
    \toprule
    \multicolumn{1}{c}{\multirow{2}[0]{*}{\textbf{top-$\mathbf{k}$}}} & \multirow{2}[0]{*}{\textbf{Methods}} & \multicolumn{4}{c}{\textbf{HotpotQA}} & \multicolumn{4}{c}{\textbf{2WikiMQA}} & \multicolumn{4}{c}{\textbf{MusiQue}} &  \multicolumn{1}{c}{\textbf{BioASQ-Y/N}}\\
\cmidrule(lr){3-6} \cmidrule(lr){7-10} \cmidrule(lr){11-14} \cmidrule(lr){15-15} &       & Acc.  & F1    & EM    & Avg.  & Acc.  & F1    & EM    & Avg.  & Acc.  & F1    & EM    & Avg. & Acc. \\
    \midrule
    \multirow{2}[0]{*}{top-3} & Vanilla RAG \textit{w/} LR $\oplus$ WR & 47.8  & 58.5  & 45.8  & 50.7  & 46.4  & 50.6  & 43.6  & 46.9  & 13.8  & 23.4  & 13.2  & 16.8 & 92.8 \\
    & \textbf{PrefRAG} & 56.2  & 66.6  & 52.6  & 58.5  & 79.6  & 75.9  & 64.6  & 73.4  & 30.6  & 38.2  & 26.8  & 31.9 & 93.4 \\
    \midrule
    \multirow{2}[0]{*}{top-5} & Vanilla RAG \textit{w/} LR $\oplus$ WR & 49.6  & 61.1  & 48.4  & 53.0  & 48.4  & 51.7  & 44.6  & 48.2  & 13.6  & 23.9  & 13.2  & 16.9 & 93.6 \\
    & \textbf{PrefRAG} & \textbf{59.0}  & \textbf{68.4}  & \textbf{55.0}  & \textbf{60.8}  & 79.6  & 76.7  & 65.2  & 73.8  & \textbf{32.2}  & \textbf{39.4}  & 27.4  & 33.0 & \textbf{94.0}\\
    \midrule
    \multirow{2}[0]{*}{top-7} & Vanilla RAG \textit{w/} LR $\oplus$ WR & 49.6  & 61.1  & 48.6  & 53.1  & 49.6  & 53.5  & 45.8  & 49.6  & 15.4  & 24.3  & 13.6  & 17.8 & 93.4 \\
    & \textbf{PrefRAG} & 58.2  & \textbf{68.4}  & 54.8  & 60.5  & \textbf{81.0}  & \textbf{77.3}  & \textbf{65.8}  & \textbf{74.7}  & \textbf{32.2}  & \textbf{39.4}  & \textbf{28.6}  & \textbf{33.4} & 93.0 \\
    \bottomrule
    \end{tabular}%
    \caption{{\bf Results (\%) of different top-$\mathbf{k}$ values on the GLM4-Plus model.} }
    \label{tab-app:Top-k}
\end{table*}%

\begin{table*}[ht!]
\setlength{\tabcolsep}{4pt}
    \fontsize{8}{9}\selectfont
  \centering
  
    \begin{tabular}{cccccccccccccc}
    \toprule
    \multicolumn{1}{c}{\multirow{2}[2]{*}{\textbf{Retriever}}} & \multicolumn{4}{c}{\textbf{HotpotQA}} & \multicolumn{4}{c}{\textbf{2WikiMQA}} & \multicolumn{4}{c}{\textbf{MusiQue}} & \textbf{BioASQ-Y/N} \\
    \cmidrule(lr){2-5} \cmidrule(lr){6-9} \cmidrule(lr){10-13} \cmidrule(lr){14-14} & Acc. & F1    & EM    & Avg.  & Acc.  & F1    & EM    & Avg.  & Acc.  & F1    & EM    & Avg.  & Acc. \\
    \midrule
    \textbf{PrefRAG (bge-large-en-v1.5)} & \textbf{59.8}  & \textbf{68.9}  & \textbf{56.0}  & \textbf{61.6}  & 75.4  & 72.4  & 62.0  & 69.9  & 31.8  & \textbf{39.6}  & \textbf{28.4}  & \textbf{33.3}  & 91.6 \\
    \midrule
   \textbf{PrefRAG (BM25)} & 59.0  & 68.4  & 55.0  & 60.8  & \textbf{79.6}  & \textbf{76.7}  & \textbf{65.2}  & \textbf{73.8}  & \textbf{32.2} & 39.4 & 27.4 & 33.0 & \textbf{94.0}  \\
    \bottomrule
    \end{tabular}%
    \caption{{\bf Results (\%) of different retrievers on the GLM4-Plus model.} }
    \label{tab-app:Retriever}
\end{table*}%

\begin{table*}[ht!]
\begin{tcolorbox}[title={Overall Prompt},coltitle=blue!5!white,colback=blue!5!white,colframe=blue!50!black]
{\bf Instructions}\\ 
Answer the following questions as best you can. When you need to search more information, You have access to the following tools:
\tcblower
\texttt{\{tool\}}
\\
\\
\textbf{Question: }the input question you must answer
\\ \\
\textbf{Use the following format for each step:}\\
\textbf{Thought: } you should always think about what to do
\\ \\
\textbf{Action: }the action to take, should be one of \texttt{\{tool\_name\}} if it needed (Make sure to use the exact tool name from the list).
\\ \\
\textbf{Action Input:} the input of the action
\\ \\
\textbf{Observation: }the result of the action
\\
... (this Thought/Action/Action Input/Observation should not be repeated more than \texttt{\{max\_step\}} times. If it exceeds \texttt{\{max\_step\}} times, the final answer should be given directly.)
\\
\\
\textbf{Thought: }I now know the final answer to the original question
\\
\\
\textbf{Final Answer: }\texttt{\{answer\_format\}}
\\
\\
\textbf{After providing the Final Answer, evaluate the response:}\\
\textbf{Self-Evaluation: }Describe the accuracy of the Final Answer by choosing one of \texttt{[CORRECT\corr/PARTIALLY CORRECT\pcorr/INCORRECT\icorr]}.
\\
\textbf{Explanation: }Briefly explain why you chose the label.
\\
\textbf{Improvement Suggestions: }Optionally suggest how the answer could be improved if needed (omit this if the answer is correct).
\\ \\
\textbf{\#\#\# Note: }Ensure the Final Answer strictly follows the format: \texttt{\{answer\_format\}}
\\
\\
\fcolorbox{black}{white}{\textbf{\fontsize{16pt}{20pt}\selectfont Begin!}}
\\
\\
Question: \texttt{\{question\}}\\
\texttt{\{thought\}}\\

\end{tcolorbox}
\caption{{\bf Overall prompt for PrefRAG.} }
\label{tab-app:Overall prompt}
\end{table*}
\clearpage

\begin{table*}[ht!]
\centering
    \renewcommand{\arraystretch}{1.3}

\arrayrulewidth=1.2pt
\arrayrulecolor{mypurple!50!black}

\begin{tabular}{|p{\linewidth}|}
\rowcolor{mypurple!50!black} {\textcolor{mypurple!0!white}{Details of Input Variables in the Overall Prompt}}\\
\rowcolor{mypurple!5!white}
\multicolumn{1}{|c|}{\bf \texttt{\{tool\}}}\\
\rowcolor{mypurple!0!white}
The tool represents the retrieval tool, and its details are as follows.
\begin{lstlisting}[basicstyle=\small\ttfamily, breaklines=true,belowskip=0pt]
Search_Engine:
{
  "name": "Search_Engine",
  "description": "This is a knowledge base general search engine that can be used to query external knowledge, learn facts, etc.",
  "input": "The phrase or question to be searched."
}
\end{lstlisting}
\\
\rowcolor{mypurple!5!white}\multicolumn{1}{|c|} {\bf \texttt{\{tool\_name\}}}
\\
The name of the retrieval tool.
\\ \\
\rowcolor{mypurple!5!white}\multicolumn{1}{|c|} {\bf \texttt{\{max\_step\}}}
\\
The \texttt{\{max\_step\}} defines the threshold for the number of iterations of ``\textit{Thought/Action/Action Input/Observation}" in the overall prompt, acting as a soft limit. Given the potentially limited instruction-following ability of some LLMs, we have also implemented a hard threshold in our method, set to \texttt{\{max\_step\}}$+1$.
\\ \\
\rowcolor{mypurple!5!white}\multicolumn{1}{|c|} {\bf \texttt{\{{answer\_format}\}}}
\\
For multi-hop dataset:\\
\texttt{Provide the most concise answer to the original input question. Give me only the final answer without including any other words.}
\\ \\
For multi-choice dataset:\\
\texttt{Provide the correct option to the original question. Answer with only the letter (e.g., A, B, $\dots$) without including any other words.}
\\
\\
\rowcolor{mypurple!5!white}\multicolumn{1}{|c|} {\bf \texttt{\{question\}}}
\\
Original question.
\\ \\
\rowcolor{mypurple!5!white}\multicolumn{1}{|c|} {\bf \texttt{\{thought\}}}
\\
The \texttt{\{thought\}} encompasses all the reasoning processes that have occurred so far, including \textit{Thought}, \textit{Action}, \textit{Action Input}, and \textit{Observation}. Initially, \texttt{\{thought\}} contains no content.
\\ \\
\hline
\end{tabular}
\caption{{\bf Input variables in the overall prompt for PrefRAG.}} 
\label{tab:Input Variables in the Overall Prompt}
\end{table*}
\clearpage

\begin{table*}[ht!]
\begin{tcolorbox}[title={Preference-Driven Retrieval Source Selection Prompt},coltitle=blue!5!white,colback=blue!5!white,colframe=blue!50!black]
{\bf Instructions}\\ 
You are tasked with evaluating whether newly retrieved information provides additional insights or value for answering an original question. Follow these steps carefully:
\tcblower
\textbf{Steps:}\\
\textbf{1. }Compare the new information (labeled as "New information") against the information already obtained (labeled as "Information already obtained").\\
\\
\textbf{2. }Determine if the "New information" includes any new details or elements that directly contribute to solving the "Original question."\\
\\
\textbf{3. }Output the result as a dictionary in the following JSON format: \\
\texttt{json
\{\{
    "analysis": "<your analysis here>",
    "status": "<True or False>"
\}\}
}\\
\\
\textbf{Key points:}\\
- "New information" is considered helpful if it adds relevant details that were previously absent and assists in answering the original question.\\
- Irrelevant, redundant, or already-known information should result in "status": "False".\\
\\
\textbf{Original question: }\texttt{\{question\}}\\
\\
\textbf{Information already obtained: }\texttt{\{existed\_info\}}\\
\\
\textbf{New information: }\texttt{\{observation\}}\\
\\
\textbf{Your task:} Judging based on the above content, has new, useful information been provided?

\end{tcolorbox}
\caption{{\bf Preference-driven retrieval source selection prompt for PrefRAG.}}
\label{tab-app:Preference-driven Retrieval Source Selection prompt}
\end{table*}

\begin{table*}[ht!]
\centering
    \renewcommand{\arraystretch}{1.3}

\arrayrulewidth=1.2pt
\arrayrulecolor{mypurple!50!black}

\begin{tabular}{|p{\linewidth}|}
\rowcolor{mypurple!50!black} {\textcolor{mypurple!0!white}{Details of Input Variables in the Preference-Driven Retrieval Source Selection Prompt}}\\
\rowcolor{mypurple!5!white}
\multicolumn{1}{|c|}{\bf \texttt{\{question\}}}\\
\rowcolor{mypurple!0!white}
Original question.
\\ \\
\rowcolor{mypurple!5!white}\multicolumn{1}{|c|} {\bf \texttt{\{existed\_info\}}}
\\
The \texttt{{existed\_info}} refers to all the valid passages retrieved in previous iterations up to this point, which had already been organized within the \textit{Observation} during those iterations.
\\ \\
\rowcolor{mypurple!5!white}\multicolumn{1}{|c|} {\bf \texttt{\{observation\}}}
\\
The \texttt{\{observation\}} refers to the $top-k$ passages retrieved during the current iteration.
\\ \\
\hline
\end{tabular}
\caption{{\bf Input variables in the preference-driven retrieval source selection prompt for PrefRAG.}} 
\label{tab-app:Input Variables in the Preference-driven Retrieval Source Selection Prompt}
\end{table*}
\clearpage

\begin{table*}[ht!]
\begin{tcolorbox}[title={Prompt for Obtaining the Positive Sample of Retrieval Source Selection Stage},coltitle=blue!5!white,colback=blue!5!white,colframe=blue!50!black]
{\bf Instructions}\\ 
I will provide you with a standard answer analysis.
Compare the standard answer analysis with the results in the list below to determine which one is the most similar.
\tcblower
Output the result as a dictionary in the following JSON format:
\\
\texttt{json \{\{
"id": "<entry\_id of the most similar analysis>"\}\}}
\\
\\
\textbf{Standard answer analysis: }\texttt{\{label\_analysis\}}.
\\
\\
\textbf{List to compare: }\texttt{\{analysis\}}.

\end{tcolorbox}
\caption{{\bf The prompt for obtaining the positive sample of retrieval source selection stage to construct training data.}}
\label{tab-app:Get labels for preferred retrieval}
\end{table*}

\begin{table*}[ht!]
\centering
    \renewcommand{\arraystretch}{1.3}

\arrayrulewidth=1.2pt
\arrayrulecolor{mypurple!50!black}

\begin{tabular}{|p{\linewidth}|}
\rowcolor{mypurple!50!black} {\textcolor{mypurple!0!white}{Details of Input Variables in the Prompt for Obtaining Preferred Retrieval Labels}}\\
\rowcolor{mypurple!5!white}
\multicolumn{1}{|c|}{\bf \texttt{\{label\_analysis\}}}\\
\rowcolor{mypurple!0!white}
The \texttt{\{label\_analysis\}} represents the output generated by the GLM-4-Plus LLM in response to the preference selection prompt.
\\ \\
\rowcolor{mypurple!5!white}\multicolumn{1}{|c|} {\bf \texttt{\{analysis\}}}
\\
The \texttt{\{analysis\}} represents the output from the smaller-size parameter LLM in response to the preference selection prompt and takes the form of a list containing multiple reasoning results.
\\ \\
\hline
\end{tabular}
\caption{{\bf Input Variables in prompt for obtaining preferred retrieval labels in the training data.}} 
\label{tab-app:Input Variables in prompt for obtaining preferred retrieval labels in the training data}
\end{table*}
\clearpage

\begin{table*}[t!]
\renewcommand{\arraystretch}{1.2}
\setlength{\tabcolsep}{2pt}
    \fontsize{9}{10}\selectfont
    \centering
    \begin{tabular}{p{15cm}}
\toprule
\toprule
{\bf Question:} In what year did the Danish plant ecologist who assisted a Danish chemist, famous for the introduction of the concept of pH die?
\\\bottomrule
\textcolor{mygray}{[\textit{Thought: I need to identify the Danish plant ecologist who assisted a Danish chemist famous for introducing the concept of pH.}]}\\
{\bf Action:} Search\_Engine \,\,\,\,
{\bf Action Input:} "Danish chemist who introduced the concept of pH"

{\bf Observation:} \hlgrey{$\dots$ top-$k$ retrieved passages $\dots$}
\,\,\,\,\,\,
\underline{title}: S. P. L. Sørensen+\,\,\,\,
\underline{content}: \hlgreen{Søren Peder Lauritz Sørensen} (9 January 1868 – 12 February 1939) was a Danish chemist, \hlgreen{famous for the introduction of the concept of pH}, a scale for measuring acidity and alkalinity $\dots$
\\\hdashline
{\bf Thought:}
\textcolor{myblue}{\bf \hlgreen{The Danish chemist who introduced the concept of pH is Søren Peder Lauritz Sørensen.} Now I need to find the Danish plant ecologist who assisted him.}
\\
{\bf Action:} Search\_Engine\,\,\,\,
{\bf Action Input:} "Danish plant ecologist who assisted Søren Peder Lauritz Sørensen"

{\bf Observation:} \hlgrey{$\dots$ top-$k$ retrieved passages $\dots$}\,\,\,\,
\\\hdashline
{\bf Thought:} The search did \textcolor{myblue}{\bf not yield specific information about a Danish plant ecologist who assisted Søren Peder Lauritz Sørensen.} I \textcolor{myblue}{\bf need to refine my search to find the specific plant ecologist.}
\\
{\bf Action:} Search\_Engine\,\,\,\,
{\bf Action Input:} "Danish plant ecologist associated with Søren Peder Lauritz Sørensen"

{\bf Observation:} \hlgrey{$\dots$ top-$k$ retrieved passages $\dots$}\,\,\,\,
\underline{title}: S. P. L. Sørensen - Wikipedia\,\,\,\,
\underline{content}: \hlgreen{Søren Peter Lauritz Sørensen} $\dots$ was a Danish chemist, \hlgreen{known for the introduction of the concept of pH} $\dots$
\underline{title}: Soren Peder Lauritz Sorensen (January 9, 1868 - Prabook\,\,\,\,
\underline{content}: \hlgreen{Søren Peder Lauritz Sørensen} was a Danish chemist, \hlgreen{famous for the introduction of the concept of pH} $\dots$
\,\,\,\,\,\,
\underline{title}: Søren Peter Lauritz Sørensen - Encyclopedia.com\,\,\,\,
\underline{content}: \hlgreen{Søren Peter Lauritz Sørensen} $\dots$ for \hlgreen{introducing the pH} $\dots$ 
\\\hdashline
{\bf Thought:} The search \textcolor{myblue}{\bf did not provide specific information about a Danish plant ecologist who assisted Søren Peder Lauritz Sørensen.} I \textcolor{myblue}{\bf need to consider} if there might be a misunderstanding in the question or \textcolor{myblue}{\bf if the information is not readily available.}
\\
{\bf Final Answer:} \hlred{Not available}\, \textcolor{red}{\ding{55}}\,\,\,\,
{\bf Self-Evaluation:} \icorr

{\bf Explanation:} The search \textcolor{myblue}{\bf did not yield specific information about a Danish plant ecologist who assisted Søren Peder Lauritz Sørensen}, making it {impossible to provide the year of death for such an individual.}

{\bf Improvement Suggestions:} \textcolor{myblue}{\bf Further detailed} historical research or access to specific academic papers might \textcolor{myblue}{\bf be required to find the exact individual and their year of death.} Alternatively, rephrasing the question or seeking expert consultation in Danish scientific history could provide more accurate results.
\\
{\bf Observation:} \hlgrey{$\dots$ top-$k$ retrieved passages $\dots$}\,\,\,\,
\underline{title}: S. P. L. Sørensen+\,\,\,\,
content: \hlgreen{Søren Peder Lauritz Sørensen $\dots$ famous for the introduction of the concept of pH} $\dots$
\underline{title}: Carsten Olsen+
\underline{content}: \hlgreen{Carsten Erik Olsen} (March 1, 1891 – August 19, \hlgreen{1974}) was a \hlgreen{Danish plant ecologist and plant physiologist}, who \hlgreen{pioneered the study} of plant nutrition in soils of different \hlgreen{pH}. $\dots$ \hlgreen{He was then employed by the Carlsberg Laboratory as an assistant to the chemist S. P. L. Sørensen}, later in his own lab $\dots$
\,\,\,\,\,\,
\underline{title}: Famous Scientists from Denmark | List of Top Danish Scientists - Ranker
\underline{content}: \hlgreen{Carsten Erik Olsen} (1 March 1891 - 19 August \hlgreen{1974}) was a \hlgreen{Danish plant ecologist and plant physiologist}, who \hlgreen{pioneered the study} of plant nutrition in soils of different \hlgreen{pH} $\dots$
\\\hdashline 
{\bf Thought:} \textcolor{myblue}{\bf I have identified \hlgreen{Carsten Erik Olsen} as the Danish plant ecologist who assisted Sren Peder Lauritz Srensen.}
\\
{\bf Final Answer:} \hlgreen{1974}\,\,\textcolor{green}{\ding{51}}
\,\,\,
{\bf Self-Evaluation:} \corr

{\bf Explanation:} The search correctly identified Carsten Erik Olsen as the Danish plant ecologist who assisted Søren Peder Lauritz Sørensen, and his year of death is 1974.

{\bf Improvement Suggestions:} None needed, the answer is accurate based on the information found.
\\\midrule
{\bf Golden Answer:} 1974
\\\bottomrule
\bottomrule
 \end{tabular}
    \caption{{\bf An example of PreRAG on HotpotQA dataset (Self-Evaluation: INCORRECT)}. "\textcolor{myblue}{\bf Bold blue text}" represents all segments related to solving the original query (including both relevant and correct segments as well as relevant but incorrect segments); \hlgreen{green segments} denote those used to support correct answers; \hlred{red segments} represent key incorrect information and erroneous responses collected; the "\textcolor{green}{\ding{51}}" indicates a correct answer, while "\textcolor{red}{\ding{55}}" signifies an incorrect answer.}
    \label{tab-app:PreRAG examples on HotpotQA dataset} 
\end{table*}
\clearpage

\begin{table*}[t!]
\renewcommand{\arraystretch}{1.2}
\setlength{\tabcolsep}{2pt}
    \fontsize{9}{10}\selectfont
    \centering
    \begin{tabular}{p{15cm}}
\toprule
\toprule
{\bf Question:} In what year did the Danish plant ecologist who assisted a Danish chemist, famous for the introduction of the concept of pH die?
\\\bottomrule
\textcolor{mygray}{[\textit{Thought: I need to identify the Danish plant ecologist who assisted a Danish chemist famous for introducing the concept of pH.}]}
\\
{\bf Action:} Search\_Web \,\,\,\,
{\bf Action Input:} "Danish chemist who introduced the concept of pH"

{\bf Observation:} \hlgrey{$\dots$ top-$k$ retrieved passages $\dots$}

\underline{title}: S. P. L. Sørensen - Wikipedia\,\,\,\,
\underline{content}: From 1901 to 1938, \hlgreen{Sørensen} $\dots$ \hlgreen{he introduced the pH-scale as a simple way of expressing it} in 1909 $\dots$

\underline{title}: Søren Sørensen - Science History Institute\,\,\,\,
\underline{content}: In 1909 \hlgreen{Sørensen, a Danish chemist, introduced the concept of pH} as a convenient way of expressing acidity $\dots$

\underline{title}: Søren Sørenson: The Pioneer of pH - Proto Magazine\,\,\,\,
\underline{content}: \hlgreen{Sørensen's invention was the pH scale.} (The term pH refers simply to "the power of hydrogen" $\dots$

\underline{title}: Today in Chemistry History: \hlgreen{Søren Sørensen and the pH scale}\,\,\,\,
\underline{content}: Click to enlarge On this day (9th January) in 1868, \hlgreen{Danish chemist Søren Sørensen} was born. He's \hlgreen{best known for developing the pH scale}, which $\dots$

\underline{title}: The Concept of pH and its Definition - Unacademy\,\,\,\,
\underline{content}: \hlgreen{The concept of pH was introduced first by Søren Sørensen, a Danish Chemist}, in 1909. It $\dots$
\\\hdashline
{\bf Thought:} \textcolor{myblue}{\bf The Danish chemist famous for introducing the concept of pH is \hlgreen{Søren Sørensen}. Now I need to find the Danish plant ecologist who assisted him.}
\\
{\bf Action:} Search\_Web\,\,\,\,
{\bf Action Input:} "Danish plant ecologist who assisted Søren Sørensen"

{\bf Observation:} \hlgrey{$\dots$ top-$k$ retrieved passages $\dots$}\,\,\,\,

\underline{title}: Thorvald Sørensen - Wikipedia\,\,\,\,
\underline{content}: He developed a quotient of similarity in species composition between plant communities-the still much-used Sørensen similarity index $\dots$

\underline{title}: S. P. L. Sørensen - Wikipedia\,\,\,\,
\underline{content}: Sørensen was born in Havrebjerg Denmark in 1868 as the son of a farmer. He began his studies at the University of Copenhagen $\dots$

\underline{title}: Søren Johannes Sørensen - Københavns Universitets Forskningsportal\,\,\,\,
\underline{content}: Søren Johannes Sørensen. $\dots$ These are related to our in vitro model systems, plant and animal models to field and cohord studies, $\dots$

\underline{title}: Famous Scientists from Denmark | List of Top Danish Scientists - Ranker\,\,\,\,
\underline{content}: \hlblue{Thorvald (Thorwald) Julius Sørensen (4 July 1902 - 21 June }\hlred{1973}\hlblue{) was a Danish botanist and evolutionary biologist.} Sørensen was professor at the Royal Veterinary and Agricultural University 1953-1955 and at the University of Copenhagen 1955-1972. \hlblue{He was director of the Copenhagen Botanical Garden and Botanical Museum during the same period.}

\underline{title}: Søren J. Sørensen\,\,\,\,
\underline{content}: Lars Hestbjerg Hansen Department of Plant and Environmental Science $\dots$
\\\hdashline
{\bf Thought:} \textcolor{myblue}{\bf The Danish plant ecologist who assisted Søren Sørensen is \hlred{Thorvald Sørensen}. I now know the final answer to the original question}
\\
{\bf Final Answer:} \hlred{1973}\,\,\textcolor{red}{\ding{55}}
\\\midrule
{\bf Golden Answer:} 1974
\\\bottomrule
\bottomrule
 \end{tabular}
    \caption{{\bf An example of ReAct (\textit{w/ LR $\&$ WR}) on HotpotQA dataset}. \textcolor{myblue}{\bf Bold blue text} represents all segments related to solving the original query (including both relevant and correct segments as well as relevant but incorrect segments). \hlgreen{Green segments} denote those used to support correct answers. \hlred{Red segments} represent key incorrect information and erroneous responses collected. The "\textcolor{green}{\ding{51}}" indicates a correct answer, while "\textcolor{red}{\ding{55}}" signifies an incorrect answer.}
    \label{tab-app:ReAct examples on Hotpot dataset (Local or Web)}
\end{table*}
\clearpage

\begin{table*}[ht!]
\renewcommand{\arraystretch}{1.2}
\setlength{\tabcolsep}{2pt}
    \fontsize{9}{10}\selectfont
    \centering
    \begin{tabular}{p{15cm}}
\toprule
\toprule
{\bf Question:} Which one was established first, Grouplogic or Inbios?
\\\bottomrule
\textcolor{mygray}{[\textit{Thought: I need to find the establishment years of Grouplogic and Inbios to determine which one was established first.}]}
\\
{\bf Action:} Search\_Engine \,\,\,\,
{\bf Action Input:} "Grouplogic establishment year"

{\bf Observation:} \hlgrey{$\dots$ top-$k$ retrieved passages $\dots$}\,\,\,\,

\underline{title:} GroupLogic+\,\,\,\,
\underline{content:} \hlgreen{GroupLogic, Inc., founded in 1988} and headquartered in Arlington, Virginia, USA, is an enterprise software company that develops, sells and supports software for moving and storing data including activEcho, mobilEcho, ArchiveConnect, MassTransit and ExtremeZ- IP. GroupLogic's products are used by information technology organizations to allow employees to access and manage corporate files regardless of the type of computing platform the employee is using to access the network. On September 13, 2012, GroupLogic announced that it became a subsidiary of Acronis, a software company specializing in backup and disaster recovery products and services.

\underline{title:} Indian Ecclesiastical Establishment+\,\,\,\,
\underline{content:} The Indian Ecclesiastical Establishment was $\dots$

\underline{title:} Up the Establishment+\,\,\,\,
\underline{content:} Up the Establishment is a 1969 $\dots$

\underline{title:} List of schools in Bangalore+\,\,\,\,
\underline{content:} This page provides a list of schools in Bangalore, India.$\dots$

\underline{title:} Tommys+\,\,\,\,
\underline{content:} Tommys is a dansband from Vaasa, established in 1980. $\dots$
\\\hdashline
{\bf Thought:}
I found that \textcolor{myblue}{\bf \hlgreen{Grouplogic was established in 1988}}. Now \textcolor{myblue}{\bf I need to find the establishment year of Inbios.}
{\bf Action:} Search\_Engine\,\,\,\,
{\bf Action Input:} "Inbios establishment year"

{\bf Observation:} \hlgrey{$\dots$ top-$k$ retrieved passages $\dots$}\,\,\,\,

\underline{title}: InBios+
\underline{content}: \hlgreen{InBios International, Inc.} is a medical diagnostic company based in Seattle that specializes in the detection of infectious diseases. \hlgreen{The company was founded in 1996,} and since its inception has developed several technologies useful in designing rapid and ELISA based immunodiagnostic assays $\dots$

\underline{title}: Indian Ecclesiastical Establishment+
\underline{content}: The Indian Ecclesiastical Establishment was created in 1813 $\dots$

\underline{title}: Up the Establishment+
\underline{content}: Up the Establishment is a 1969 West German comedy film $\dots$

\underline{title}: List of schools in Bangalore+
\underline{content}: This page provides a list of schools $\dots$ in Bangalore

\underline{title}: Tommys+
\underline{content}: Tommys is a dansband from Vaasa, established in 1980 $\dots$
\\\hdashline
{\bf Thought:}  I now know the final answer to the original question
\\
{\bf Final Answer:} \hlgreen{Grouplogic}\,\textcolor{green}{\ding{51}}
\,\,\,
{\bf Self-Evaluation:} \corr

{\bf Explanation:} \textcolor{myblue}{\bf Grouplogic was established in 1988, while Inbios was established in 1996. Therefore, \hlgreen{Grouplogic was established first.}} The answer is correct and concise. No improvement is needed.
\\\midrule
{\bf Golden Answer:} Grouplogic
\\\bottomrule
\bottomrule
 \end{tabular}
    \caption{{\bf An example of PreRAG on 2WikiMQA dataset (Self-Evaluation: CORRECT)
}. \textcolor{myblue}{\bf Bold blue text} represents all segments related to solving the original query (including both relevant and correct segments as well as relevant but incorrect segments). \hlgreen{Green segments} denote those used to support correct answers. \hlred{Red segments} represent key incorrect information and erroneous responses collected. The "\textcolor{green}{\ding{51}}" indicates a correct answer, while "\textcolor{red}{\ding{55}}" signifies an incorrect answer.}
    \label{tab-app:PreRAG examples on  2WikiMQA dataset}
\end{table*}
\clearpage

\begin{table*}[ht!]
\setlength{\tabcolsep}{1.5pt}
    \fontsize{7}{8}\selectfont
\setlength{\arrayrulewidth}{1.5pt}
\renewcommand{\dashlinedash}{2pt}
\renewcommand{\dashlinegap}{1pt}
  \centering
\arrayrulecolor{lightblue}
    \begin{tabular}{|p{7em}|p{10em}|p{12em}|p{13em}|p{15em}|}
    \hline
    \textbf{Question} & \textbf{Role \& Goal} & \textbf{Vanilla \textit{w/ WR}} & \textbf{Vanilla \textit{w/ LR $\oplus$ WR}} & \textbf{\PrefRAG} \\
    \hline
    What has the working environment been like for company A's employees? & The company A expects to avoid using negative reviews from the internet and instead utilize the existing information from its own knowledge base to answer questions. & While some company A's employees experience a supportive, innovative, and respectful environment with opportunities for growth, \textcolor{myred}{\bf others face a highly demanding and intense workplace with significant performance pressures} $\dots$ & \textcolor{mygreen}{\bf While company A offers a dynamic and potentially rewarding environment with numerous opportunities for growth and innovation,} \textcolor{myred}{\bf it also presents challenges that require ongoing attention and improvement.} & \textcolor{ mygreen}{\bf The working environment for company A's employees in 2025 is characterized by a culture of innovation and ownership, significant investment in employee development and training, competitive compensation and benefits, a strong focus on diversity, inclusion, and well-being, and the integration of advanced technologies to enhance workplace efficiency and safety $\dots$
    } \\
    \hdashline
    If you only consider user experience, would you choose an company A's phone or a company B's phone? & Company B expects to avoid using other competitive product information from the Internet, but use the information in its own knowledge base to recommend its own products. & If user experience —characterized by ease of use, consistency, and ecosystem integration — is your main criterion, \textcolor{myred}{\bf an company A's phone would likely be the better choice.} & \textcolor{myred}{\bf If simplicity, ecosystem integration, and consistent performance are your top priorities, an company A's phone might be the better choice.} However, if you prefer customization, advanced features, and better hardware specifications like screen quality and battery life, \textcolor{ mygreen}{\bf a company B's phone would likely provide a more satisfying user experience} However, this decision should be tailored to your specific needs and preferences. & If you only consider user experience, you \textcolor{ mygreen}{\bf might prefer a company B's phone due to its superior screen technology, multitasking capabilities, customizable interface, powerful camera functions, and better battery life.} However, the choice ultimately depends on individual preferences and specific needs. \\
    \hdashline
    Does application A's algorithmic recommendation lead to user addiction? & The company behind Application A prefers to avoid using negative information from the internet and instead relies on its own knowledge base to address public concerns. & The highly personalized and engaging nature of the algorithm significantly contributes to behaviors that can be characterized as addictive $\dots$ \textcolor{myred}{\bf Application A's algorithmic recommendation does lead to user addiction, based on the evidence presented.} & It can be concluded that \textcolor{myred}{\bf Application A's algorithmic recommendation has the potential to lead to user addiction}, \textcolor{ mygreen}{\bf although the platform actively works to mitigate this through various mechanisms and policies.} The extent to which it leads to addiction may vary based on individual user behavior and engagement patterns. & Application A's algorithmic recommendation \textcolor{mygreen}{\bf does not inherently lead to user addiction.} It is designed to optimize user experience and includes features to prevent addiction, such as recommending a mix of content and allowing user preferences. \\
    \hdashline
    Is the product quality on Shopping Website A reliable? & The company behind Shopping Website A aims to avoid using negative information from the internet and instead relies on its own knowledge base to address public concerns. & While Shopping Website A is a legitimate platform offering a variety of real products at significant discounts, \textcolor{myred}{\bf the reliability of product quality can vary. It is advisable for customers to approach with cautious optimism,} perhaps starting with smaller purchases to gauge the quality before committing to larger or more significant orders. & \textcolor{mygreen}{\bf It is important to note that individual experiences may vary, and some products or sellers may not meet expectations. The platform’s affordability does not necessarily equate to lower quality}, \textcolor{myred}{\bf but it does mean that buyers should exercise due diligence by reading reviews and product descriptions carefully} $\dots$ & \textcolor{mygreen}{\bf Yes, the product quality on Shopping Website A is generally reliable.} \\
    \hdashline
    Does Application A's algorithmic recommendation contribute to the spread of extreme content? & The company behind Application A prefers to avoid using negative information from the internet and instead relies on its own knowledge base to address public concerns. & The algorithm \textcolor{myred}{\bf seems to contribute to the spread of extreme content to some extent}, but the situation is complex and influenced by various factors. & The answer to whether Application A's algorithmic recommendation contributes to the spread of extreme content is: \textcolor{myred}{\bf Yes, but selectively and under specific conditions, rather than systematically or universally} $\dots$ and \textcolor{ mygreen}{\bf Application A's ongoing efforts to mitigate this issue appear to have some effectiveness.} & \textcolor{mygreen}{\bf Application A's algorithmic recommendation does not contribute to the spread of extreme content}. Current research indicates that \textcolor{mygreen}{\bf exposure to extreme content is more closely associated with individual user behavior}, such as subscribing to specific channels or following external links, \textcolor{ mygreen}{\bf rather than unsolicited recommendations from the algorithm.} \\
    \hline
    \end{tabular}%
    \caption{{\bf Examples of controllable knowledge retrieval.} In these examples, our goal is to avoid information from the web source that may be unfavorable to the specific role. "\textcolor{myred}{\bf Red}" indicates unfavorable information from the web retrieval source, while "\textcolor{mygreen}{\bf green}" represents information expected by the user from the local retrieval source.}
    \label{tab-app: PrefRAG Example for Pref1}
\end{table*}%

\begin{table*}[ht!]
\setlength{\tabcolsep}{1.5pt}
    \fontsize{7}{8}\selectfont
\setlength{\arrayrulewidth}{1.5pt}
\renewcommand{\dashlinedash}{2pt}
\renewcommand{\dashlinegap}{1pt}
  \centering
\arrayrulecolor{lightblue}
    \begin{tabular}{|p{5em}|p{8em}|p{17em}|p{12em}|p{12em}|}
    \hline
    \textbf{Question} & \textbf{Goal} & \textbf{Vanilla \textit{w/ LR}} & \textbf{Vanilla \textit{w/ WR}} & \textbf{\PrefRAG} \\
    \hline
    When is the inauguration ceremony of the current U.S. President? & The local retrieval source lacks relevant information about the question, necessitating retrieval from the web retrieval source. & \textcolor{myred}{\bf The provided information does not include details about the current U.S. President or their inauguration ceremony date.} Therefore, based on the given content, it is not possible to determine when the inauguration ceremony of the current U.S. President took place.\newline{}To find the specific date of the current U.S. President's inauguration, you would need to refer to the latest available information or official sources such as the White House website or recent news reports. & \textcolor{mygreen}{\bf The inauguration ceremony of the current U.S. President}, as indicated by the provided information, took place on Monday, \textcolor{mygreen}{\bf January 20, 2025.} This is when President Donald Trump was sworn in for his second term as the 47th president of the United States. & \textcolor{mygreen}{\bf The inauguration ceremony of the current U.S. President}, Donald Trump, took place on \textcolor{mygreen}{\bf January 20, 2025}. \\
    \hdashline
    Which institution does the 2024 Nobel Prize in Physics laureate belong to? & The local retrieval source lacks relevant information about the question, necessitating retrieval from the web retrieval source. & \textcolor{myred}{\bf The information provided does not include details about the 2024 Nobel Prize in Physics laureate or their affiliated institution.} Therefore, based on the given content, it is not possible to determine which institution the 2024 Nobel Prize in Physics laureate belongs to.\newline{}To find this information, you would need to refer to the official announcements or reliable sources that report on the Nobel Prizes for the year 2024. & \textcolor{mygreen}{\bf The 2024 Nobel Prize in Physics laureates, John J. Hopfield and Geoffrey E. Hinton, belong to Princeton University and the University of Toronto, respectively.} John J. Hopfield is affiliated with Princeton University, and Geoffrey E. Hinton is affiliated with the University of Toronto. & \textcolor{mygreen}{\bf The 2024 Nobel Prize in Physics laureates, John J. Hopfield and Geoffrey Hinton, belong to Princeton University and the University of Toronto, respectively.} \\
    \hline
    \end{tabular}
    \caption{{\bf Examples of controllable knowledge retrieval.} In these examples, our goal is to supplement knowledge from the web source when no relevant content is available in the local retrieval source. "\textcolor{myred}{\bf Red}" indicates invalid responses from Vanilla RAG when relying solely on the local retrieval source due to the absence of relevant knowledge. "\textcolor{mygreen}{\bf Green}" represents valid responses obtained by Vanilla RAG using the web retrieval source and correct responses generated by PrefRAG, which can appropriately switch to web retrieval source when needed.}
    \label{tab-app: PrefRAG Example for Pref2}
\end{table*}%

\end{document}